\documentclass[journal]{IEEEtran}
\usepackage{xcolor}

\usepackage{mathrsfs}
\usepackage{bm}
\usepackage{bbm}
\usepackage{upgreek}
\usepackage{booktabs}
\usepackage[T1]{fontenc}

\usepackage{ntheorem}

\newtheorem*{definition-non}{Definition}

\usepackage{threeparttable}

\usepackage{lipsum}

\usepackage{cite}
\ifCLASSINFOpdf
   \usepackage[pdftex]{graphicx}
\else
   \usepackage[dvips]{graphicx}
\fi
\usepackage{amsmath}
\usepackage{algorithmic}
\usepackage{array}
\ifCLASSOPTIONcompsoc
  \usepackage[caption=false,font=normalsize,labelfont=sf,textfont=sf]{subfig}
\else
  \usepackage[caption=false,font=footnotesize]{subfig}
\fi
\usepackage{fixltx2e}
\usepackage{placeins}

\usepackage{stfloats}
\usepackage{url}
\begin{document}
\title{Painless step size adaptation for SGD}

\author{Ilona~Kulikovskikh~and
        Tarzan~Legovi\'c
    % <-this % stops a space
\thanks{I. Kulikovskikh is with the Department
of Information Systems and Technologies, Samara University, Samara
443086, Russia (e-mail: kulikoskikh.im@ssau.ru).}% <-this % stops a space
\thanks{T.~Legovi\'c is with the Institute of Applied Ecology, Oikon Ltd., with Libertas International University, and with Ru\dj er Bo\v{s}kovi\'c Institute, Zagreb 10000, Croatia (email: tlegovic@oikon.hr)}% <-this % stops a space
\thanks{Manuscript received xx xx, xxxx; revised xx xx, xxxx. (Corresponding author: Ilona Kulikovskikh).}}


\markboth{IEEE TRANSACTIONS ON NEURAL NETWORKS AND LEARNING SYSTEMS}%
{Kulikovskikh \MakeLowercase{\textit{et al.}}: Painless step size adaptation}



\maketitle

\begin{abstract}
Convergence and generalization are two crucial aspects of performance in neural networks. When analyzed separately, these properties may lead to contradictory results. Optimizing a convergence rate yields fast training, but does not guarantee the best generalization error. To avoid the conflict, recent studies suggest adopting a moderately large step size for optimizers, but the added value on the performance remains unclear. We propose the LIGHT function with the four configurations which regulate explicitly an improvement in convergence and generalization on testing. 
This contribution allows to: 1) improve both convergence and generalization of neural networks with no need to guarantee their stability; 2) build more reliable and explainable network architectures with no need for overparameterization. We refer to it as ``painless'' step size adaptation.

\end{abstract}


% \fi
\IEEEpeerreviewmaketitle

\section{Introduction}

\IEEEPARstart{N}{eural} networks imitate signal transmission within neurons in the brain with units which are interconnected through weighted links and assembled in layers \cite{markus2001, goodfellow2016}. Training a neural network implies updating the model weights to best map inputs to outputs. This process is framed as an optimization problem that involves minimizing the model errors on a training dataset.
When training a network with gradient-based methods, accelerating convergence to the solution is of a high priority \cite{dieuleveut2017, arora2019}, but not the only performance variable to optimize. Minimizing the difference between the model errors on a training and a testing dataset, which is called the generalization error, plays a fundamental role \cite{amari1967, tang1992, hardt2016, neyshabur2017, zhan2017, giryes2018, lin2019}.    

The iterative optimization schemes with an adaptive step size schedule converge faster \cite{duchi2011, zeiler2012, kingma2015, wilson2017}, but generalize poorly  \cite{luo2019, liang2020, xie2020, heo2021, zhou2021}. They are often outperformed by non-adaptive stochastic gradient descent (SGD) \cite{robbins1951} for over-parameterized neural networks, where the number of trainable parameters is much higher than the number of samples they are trained on.
Exploring critical generalization capacity, several studies explained this phenomenon:  overparametrization ensures faster convergence \cite{arora2018, li2018, allen-zhu2019, oymak2019, liu2020, oymak2020, chen2021} while inducing implicit regularization of the original problem, which can potentially ease the minimization of the generalization error \cite{neyshabur2014, soudry2018, nacson2019, chizat2020, smith2021, wu2021, yun2021}. However, oveparameterized models require an enormous number of units and layers to represent, process, and store data. This heavily reduces the transparency of neural networks, making them difficult to interpret.

What makes neural networks generalize well? 
Relying on the extensive empirical studies of SGD, it became evident that the step size maximizing the test accuracy is usually larger than the step size which minimizes the training loss \cite{debortoli2020, li2020, cohen2021}. The occurrence of an implicit regularizer demystifies this matter as well. For a small step size, SGD behaves similar to GD on the full batch loss function. When a step size increases,  the regularizer starts penalizing the mean Euclidean norm of the minibatch gradients \cite{cohen2021, smith2021, wu2021} that makes the training loss non-monotonic. 
Another explanation is that faster gradient descent methods naturally generate chaotic dynamical systems \cite{doel2012}, which bring the optimizer to the edge of stability \cite{debortoli2020, cohen2021} and, thus, yield a non-monotone decrease pattern in the loss function. This finding shares some similarities with the edge of chaos concept \cite{schoenholz2017}. According to the concept, deep networks may be trained only sufficiently close to criticality, avoiding the regions of vanishing and exploding gradients, which correspond to the ordered and chaotic phases, respectively.  

Answering the research question, we hypothesize that non-monotonicity of the loss function itself may establish faster convergence and better generalization bounds, even without adaptive optimization tricks.  
Our reasoning comes from the thorough review of literature on the positive effect of non-monotonicity \cite{defelice1993, nie2015} in neural networks  caused by discontinuity \cite{findlay1989, wu2009, Peng2019}, delays \cite{baldi1994, lu2006, liu2012, nie2012, duan2014, nie2015}, differential inclusions, \cite{cai2012, romero2020}, sliding modes \cite{levant2013, duan2014} on the global finite-time convergence and stability \cite{forti2003, huang2008, guo2009, qin2009}. 

\begin{figure}[h!]
	\centering
	\includegraphics[width=1\columnwidth]{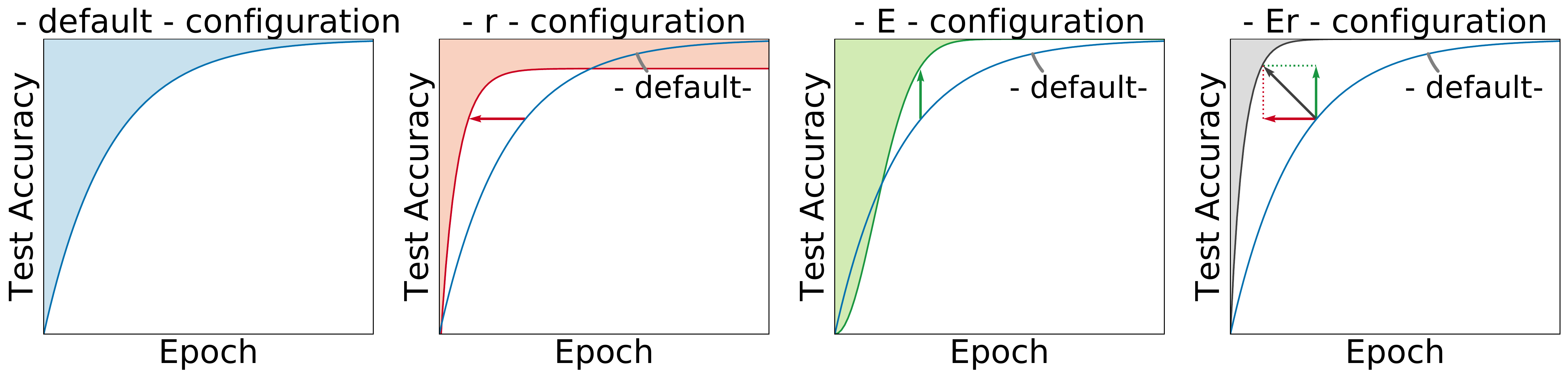}
	\caption{The balance between convergence and generalization for the LIGHT configurations}
	\label{fig:config} 
\end{figure}

To examine the hypothesis, we simulate non-monotonocity in neural networks with the LIGHT (\textbf{L}og\textbf{I}stic \textbf{G}rowth with \textbf{H}arves\textbf{T}ing) activation function, which originates from population dynamics \cite{legovic2016, gray2017} and behaves as follows. It starts growing with the rate $r$ by the logistic law. At the time $t=T$, it starts declining with the rate $E$. The y-intercepts at the moments $0$ and $T$ are specified. The default step size of SGD is modified with regard to $r$ and $E$, respectively.
For a diagnostic purpose, we suggest four configurations of the function to regulate explicitly an improvement in convergence and generalization on testing (see Fig. \ref{fig:config}):
\begin{itemize}
	\item[] -default-: no improvement;
	\item[] -r-: an improvement in convergence; 
	\item[] -E-: an improvement in generalization;
	\item[] -Er-: both convergence and generalization are improved. 
\end{itemize}

Increasing the rate $r$ allows us to push a learning system towards the edge of stability. By increasing the rate $E$, we fix this edge by pushing the system towards the equilibrium point.  The presence of sliding modes along discontinuity surfaces, which are modelled with $T$, establishes the control over the system behavior.

We show that the formulated hypothesis is valid and that the LIGHT function contributes to: 
1) improving convergence and generalization by training neural networks with a moderately large step size with no need to ensure its stability;
2) building more reliable and explainable network architectures with no need for overparameterization.
We refer to it as ``painless'' step size adaptation.

\section{Related work}

\subsection{Adaptive optimization}
SGD is one of the most dominant first-order optimization algorithms for training neural networks \cite{robbins1951}. 
Albeit its popularity and simplicity, SGD scales the gradient equally in all directions that results in worse convergence than the adaptive methods, for example, Adam \cite{kingma2015} and Adagrad \cite{duchi2011} that scale the gradient using the information from past gradients \cite{luo2019}. Adopting optimization methods with variable step sizes leads to faster convergence and worse generalization compared to non-adaptive methods \cite{hoffer2017, wilson2017, gunasekar2018}: they train faster, but their performance plateaus on testing due to less stable and predictable behavior.
Consequently, further development of optimization techniques is directed towards a better trade-off between convergence and generalization.

In order to address the above problem, the first group of studies is directed towards different SGD modifications such as
SGD with adopting extended differentiators \cite{yang2018}, 
random reshuffling \cite{ying2019},
local changes in gradients \cite{dubey2020, wang2020} and etc.

The analysis of SGD based optimization for overparameterized models has recently become another active area of research interest \cite{arora2018, li2020, arora2019, allen-zhu2019, zhou2021, wu2021}.
Recent studies indicated that large step sizes can preserve good generalization and accelerate SGD convergence without any additional gradient scaling. 
While analyzing the effect of overparametrization, Wu et al. \cite{wu2021} pointed to the difference in directional biases for SGD and GD with a moderate and annealing step size.
Vaswani et al. \cite{vaswani2019} explored line-search techniques and provided heuristics to automatically set larger step sizes.
Li and Arora \cite{li2020} carried out an analysis of an exponential step size schedule. They showed that using SGD with momentum \cite{liu2020} and an exponentially increasing step size, coupled with batch normalization, maintains good balance between convergence and generalization across all standard architectures. 
Nitanda and Suzuki \cite{nitanda2021} provided an analysis of a convergence rate for the averaged SGD in the Neural Tangent Kernel Regime. The authors disclosed the conditions on which the method can achieve the minimax optimal convergence rate, with the global convergence guarantee. 

In parallel with more successful SGD adoption for overparameterized models, substantial progress has been achieved in optimization methods with adaptive step sizes. 
SGDP and AdamP use effective step sizes without changing the update directions \cite{heo2021}. This allows to preserve the original convergence properties of GD optimizers. RAdam adopts the learning rate warmup heuristic to rectify the variance of adaptive step sizes \cite{liu2020} and, by that, stabilize training, accelerate convergence, and improve generalization. 
In an attempt to balance generalization and convergence on unstable and extreme step sizes, Luo at al. \cite{luo2019} put forward AdaBound and AMSBound which adopt dynamic bounds on step sizes to eliminate the generalization gap between adaptive methods and SGD and maintain higher learning rate early in the training. These methods were further developed with regard to a dynamic decay rate in \cite{liang2020}.
Xie at al. \cite{xie2020} proved that the normalized Adagrad ensures robustness to the choice of hyper-parameters and achieves a linear convergence rate for a subset of either strongly convex functions or non-convex functions that satisfy the Polyak-Lojasiewicz (PL) inequality.
Zhou at al. \cite{zhou2021} proposed the SAGD method that leverages differential privacy to boost the generalization performance of adaptive gradient methods.

%modification of SGD
%overparamterization and huge step sizes
% adaptive methods with better generalization
%
%1) automatic self-stability;\\
%
%simple easy-to-use approach\\
%explainable (pop dynamics)\\
%four configurations allows to regulate an improvement in convergence and generalization on testing explicitly

% learning rates (unclear what to choose, the edge of stability)

\subsection{Adaptive activation}
% activation functions (adaptive + non-monotone)
Using adaptive activation functions in neural networks is one more way to balance convergence and generalization. 
The first activation function presented ``all-or-none'' character of nervous activity with a step function \cite{mcculloch1943, rosenblatt1958} to solve a binary classification problem. 
Wilson and Cowan \cite{wilson1972} derived coupled nonlinear differential equations from the dynamics of spatially localized populations containing both excitatory and inhibitory model neurons. They investigated population responses to various types of stimuli and introduced an s-shaped monotonic function of stimulus intensity - a sigmoid function.
Yamada and Yabuta \cite{yamada1992} suggested an approach to optimally tune the shape of the sigmoid function in control systems.
While comparing the approximation capabilities of activation functions, DasGupta and Schnitger \cite{dasgupta1993} pinpointed that the standard sigmoid is more powerful than the binary threshold even when computing boolean functions \cite{sontag1989}. 
Piazza et al. \cite{piazza1993} proposed the adaptive polynomial activation function to address the issue of complexity in neural networks. 
Xu and Zhang \cite{xu2001} proposed another adaptive activation function to reduce a network size.  
Goh and Mandic \cite{goh2003} suggested to adapt the amplitude of activation functions, while reconsidering recurrent neural networks in terms of nonlinear adaptive filters.
Bai et al. \cite {bai2009} showed  that varying the slope of an activation function with different step sizes is more beneficial than using momentum and an adaptive step size in the backpropagation algorithm. 
Flennerhag \cite{flennerhag2018} suggested simple drop-in replacements that learn to adapt their parameterization with regard to the network inputs.
PPolyNets \cite{wu2018} are accurate and efficient parametric polynomial activations specifically developed for encryption schemes which support only polynomial operations.
Goyal et al. \cite{goyal2020} suggested to normalize polynomial activations to increase the stability of neural networks.
Kunc and Kl\v{e}ma \cite{kunc2021} proposed a novel transformative adaptive activation function that improves the gene expression inference by generalizing existing adaptive activation functions.

%D.A. Findlay \cite{findlay1989} considered aspects of training networks whose neurons may have discontinuous or non-differentiable activation functions. 
De Felice et al. \cite{defelice1993} drew attention to the fact that the biological activation function has a more complicated behavior which reduces to the usual (step or sigmoid) function for some hyperparameters describing its shape and stated that the non-monotonicity of the function increases the capacity of neural networks.
Baldi and Atiya \cite{baldi1994} extended previously known results regarding the effects of delays on stability and convergence properties. 
Forti and Nistri \cite{forti2003} introduced a general class of neural networks, where the neuron activations are modeled by discontinuous functions. The authors discovered that the presence of sliding modes ensures global convergence in neural networks in finite time. Duan et al. \cite{duan2014} established the existence and global exponential stability of almost periodic solution for the delayed high-order Hopfield neural networks. 
The study \cite{lu2006} discussed the dynamics of a class of the delayed neural networks with discontinuous activation functions. The authors concluded that the solution of delayed neural networks with discontinuous activation functions can be regarded as a limit of the solutions of delayed neural networks with high-slope continuous activation functions.
According to \cite{hahnloser2000, glorot2011}, rectified linear units (ReLU) and their different modifications \cite{clevert2016, zhu2021} in the hidden layers of neural networks demonstrate better convergence and generalization in comparison with the continuous activations.  
Exploring monostability and multistability of almost-periodic solutions in the fractional-order neural networks, Wan et al. \cite{wan2020} indicated 
that the dynamics in neural networks with the unsaturating piecewise linear activation functions is more complex. 
Nie and Zheng \cite{nie2015} looked into the problem of coexistence and dynamical behaviors of multiple equilibrium points for neural networks with discontinuous non-monotonic piecewise linear activation functions and time-varying delays. The study revealed that discontinuous neural networks can have greater storage capacity than the continuous ones.  
Hayou et al. \cite{hayou2019}, however, mentioned that only a specific choice of hyperparameters such as initialization and activation with regard to the concept of chaos \cite{schoenholz2017} improves convergence and generalization in neural networks.

\section{Preliminaries}
For a dataset $\{\mathrm{x}_i,y_i\}_{i=1}^m$ with $\mathrm{x}_i\in\mathrm{R}^n$, $y_i\in\{-1,1\}$, we minimize an empirical loss function for each mini-batch dataset $B(t)\subseteq \{1, \dots, m\}$ with a weight vector $\bm{\uptheta}\in\mathrm{R}^n$:
\begin{equation}
\label{eq::01}
\mathcal{L} (\Theta) = \sum_{i\in B(t)}\ell (y_i \left\langle \Pi(\Theta),\mathrm{x}_i\right\rangle),
\end{equation}
where $\ell$ measures the discrepancy between the output $y$ and the model prediction. The SGD optimizer finds the weight vector
with a fixed step size $\eta$:
\begin{equation}
\label{eq::02}
\bm{\Uptheta}_l(t+1) = \bm{\Uptheta}_l(t)-\eta\nabla_{\uptheta_l}\mathcal{L} (\Theta), 
\end{equation}
where  $\Pi(\Theta) = \bm{\Uptheta}_1\times\bm{\Uptheta}_2\times\dots\times\bm{\Uptheta}_L$, 
$\Theta = \{\bm{\Uptheta}_l\in\mathrm{R}^{d_{l-1}\times d_l}: l = 1, 2,  \dots, L\}$, $L$ is the number of layers, $d_l$ is the number of nodes in the layer $l$.

%We are particularly interested in modeling \textcolor{blue}{the loss/activation function} $\ell$ of the final classification layer with population dynamics, where the sigmoid function is currently the only option for the network binary output.

\section{LIGHT}

%Let us introduce the LIGHT function $\ell^{\hspace{1mm}r, E\hspace{1mm}} (t)$ which starts growing with the rate $r$ by the logistic law. After the time $T$, it declines with the rate $E$.
%We scaled the function so that
%$$\lim_{t\rightarrow-\infty}\ell^{\hspace{1mm}r, E\hspace{1mm}} (t)= 0, \hspace{1mm} 
%\lim_{t\rightarrow\infty}\ell^{\hspace{1mm}r, E\hspace{1mm}}(t)= \epsilon$$
%with the y-intercepts  $\ell^{\hspace{1mm}r, E} (0)= N_0$ and $\ell^{\hspace{1mm}r, E\hspace{1mm}} (T)= N_T$ and built it on the standard sigmoid.

We built a diagnostic function $\ell^{\hspace{1mm}r, E\hspace{1mm}} (t)$ on the standard sigmoid by simulating different types of non-monotonocity with the growth rate $r$ and decline rate $E$. The function grows with a constant rate $r$ according to the logistic law. After time $T$, it declines with a constant rate $E$. 
We call the function LIGHT (\textbf{L}og\textbf{I}stic \textbf{G}rowth with \textbf{H}arves\textbf{T}ing) as its behavior inherits the principles of population dynamics \cite{legovic2016, gray2017}. Let us present the LIGHT function. 

\begin{definition-non}[LIGHT]
\label{def::1}
    For any time $t\in\mathrm{R}$, time instant $T>0$, growth rate $r>0$ and decline rate $E \geq 0$, a non-monotonic function $\ell^{\hspace{1mm}r, E\hspace{1mm}} (t)$, 
    such that $$\lim_{t\rightarrow-\infty}\ell^{\hspace{1mm}r, E\hspace{1mm}} (t)= 0, \hspace{1mm} 
    \lim_{t\rightarrow\infty}\ell^{\hspace{1mm}r, E\hspace{1mm}}(t)= \varepsilon, $$ 
    where $\varepsilon$ is the extent to which $r$ is impacted by $E$, \\behaves as:
	\begin{equation}
	\label{eq::light}
	\ell^{\hspace{1mm}r, E\hspace{1mm}} (t) = 
	\begin{cases}
	\varepsilon \mathrm{e}^{\left(\ln_\emph{q}N_T+\frac{E}{r}\right)\mathrm{e}^{-r(t-T)}} \hspace{4mm}\text{if  $t \geq T$},\\
	\varepsilon \mathrm{e}^{\left(\ln_\emph{q}N_0\right)\mathrm{e}^{-rt}} \hspace{13mm}\text{otherwise},
	\end{cases}
	\end{equation}
with the derivative $\ell^{\prime\hspace{1mm}r, E\hspace{1mm}} (t)$:
	\begin{multline}
	\begin{cases}
	-\varepsilon r\left(\ln_\emph{q}N_T+\frac{E}{r}\right)\mathrm{e}^{-r(t-T)+\left(\ln_\emph{q}N_T+\frac{E}{r}\right)\mathrm{e}^{-rt}}\hspace{4mm}\text{if  $t \geq T$},\\
	-\varepsilon r\ln_\emph{q}N_0\mathrm{e}^{-rt+\ln_\emph{q}N_0\mathrm{e}^{-rt}} \hspace{28mm}\text{otherwise},
	\end{cases}
	\nonumber
	\end{multline}
where $N_0 = \ell^{\hspace{1mm}r, E} (0)$, $N_T = \ell^{\hspace{1mm}r, E\hspace{1mm}} (T)$, $\ln_\emph{q}(x)$ is the \emph{q}-logarithm, where \emph{q} is the rate with which the function grows when smaller.
\end{definition-non}
By introducing $q$, we move from the infinitesimal calculus to quantum calculus~\cite{ernst2003, jackson1908, tsallis1988,  tsallis1994} to avoid the concept of limits and, thus, simplify the definition.
Another justification of this parameter is from the point of explainability: it generalizes 
the Verhulst ($q=1$, \textbf{light-v}) \cite{verhulst1838, legovic2016, schaefer1954} and Gompertz ($q\rightarrow 0$, \textbf{light-g}) \cite{gompertz1825,tjorve2017,winsor1932} laws of population dynamics \cite[see Section 7.2.3]{gray2017}.

\begin{figure}[h!]
	\centering
	\includegraphics[width=\columnwidth]{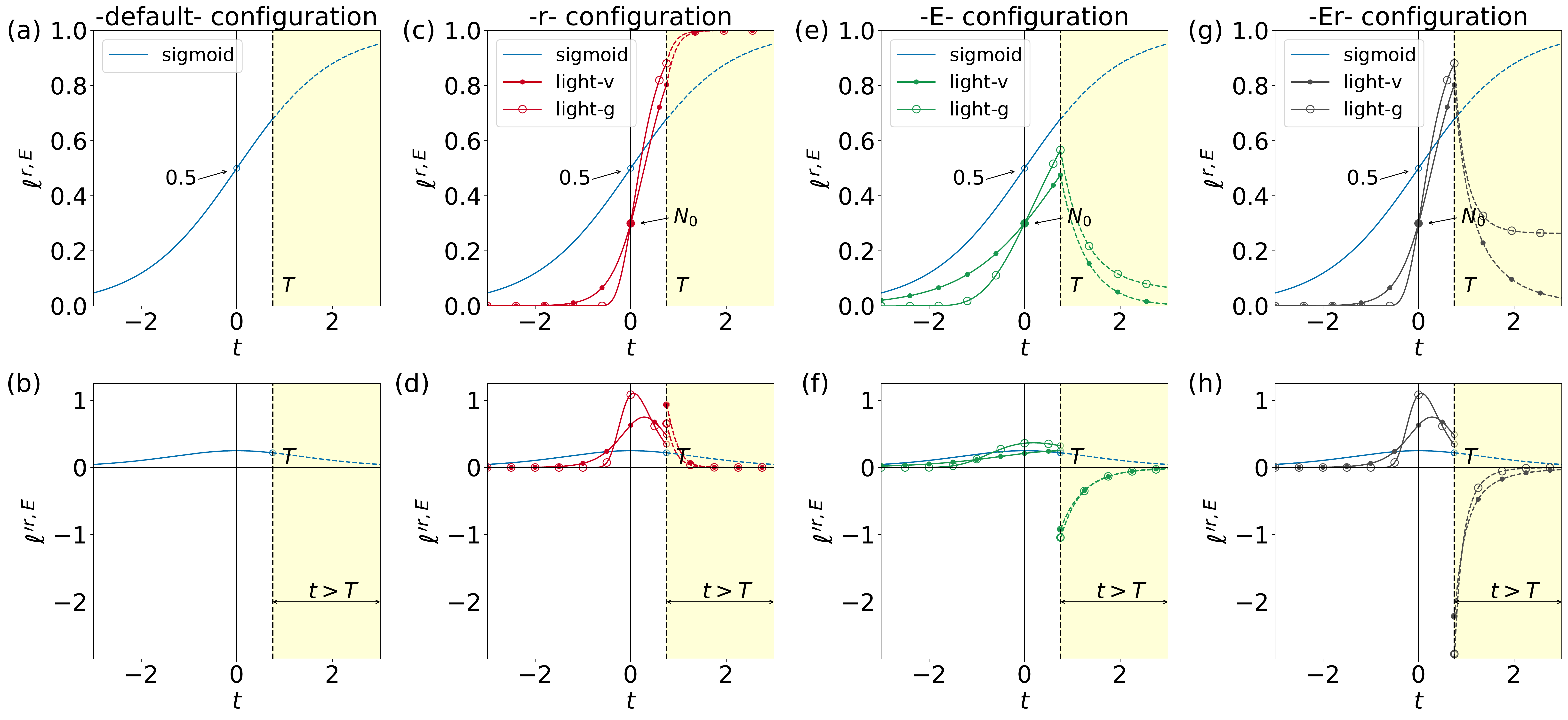}
	\caption{The LIGHT configurations and their derivatives subject to $T = 0.75$. 
		(a), (b) -default- configuration: $r = 1$, $E = 0$, $N_0 = 0.5$ (in blue).
		(c), (d) -r- configuration: $r = 3$, $E = 0$, $N_0 = 0.3$ (in red).
		(e), (f) -E- configuration: $r = 1$, $E = 3$, $N_0 = 0.3$ (in green).
		(g), (h) -Er- configuration: $r = 3$, $E = 4$, $N_0 = 0.3$ (in black).
		%(a), (b) Some examples from CIFAR-10 \cite{4}. The objects in     
		%single-label images are usually roughly aligned.(c),(d) However, the 
		%assumption of object alignment is not valid for multi-label
		%images. Also note the partial visibility and occlusion
		%between objects in the multi-label images.
	}
	\label{fig:light_config} 
\end{figure}

%The detailed derivation of the function \eqref{eq::light} is given in Appendix~\ref{proof::def}. 
To introduce non-monotonocity in the LIGHT diagnostic function, 
we propose four different configurations (see Fig. \ref{fig:light_config}). We augmented the -default- configuration, which reduces to \textbf{sigmoid} if $q=1$ and $N_0=0.5$, with three more configurations which regulate an improvement in convergence and generalization:
\begin{itemize}
	\item[] \textbf{-default-}: $r = 1$, $E=0$; no improvement;
	\item[] \textbf{-r-}:  $r \uparrow$, $E = 0$; an improvement in convergence; 
	\item[] \textbf{-E-}: $r = 1$, $E \uparrow$; an improvement in generalization;
	\item[] \textbf{-Er-}: $r \uparrow$, $E \uparrow$; an improvement in both convergence and generalization. 
\end{itemize}
The sign $\uparrow$ means that value of the parameter is significantly increased. The decline in growth, delayed by $T$, induces a discontinuity in the function. A simultaneous increase in $r$ and $E$ makes the discontinuity even more noticable by scaling the function magnitude that results in a greater impact on a convergence/generalization trade-off. 
To diagnose a neural network capability with the LIGHT function, we modified a step size $\eta$ in the SGD optimizer \eqref{eq::02} by replacing $\ell(t)$ with $\ell^{\hspace{1mm}r, E\hspace{1mm}} (t)$ in the loss \eqref{eq::01}.

\section{Experimental Results}

\subsection{Experimental Setup}
As noted in the surveyed literature on adaptive optimization and activation, the success in balancing convergence and generalization is often attributed to the complexity and capacity of neural networks. 
To clearly observe the distinctive contribution of the LIGHT function to the step size adaptation, we focused primarily on creating the simplest network architecture - a single neuron $L=0$ \cite{minsky1969, aizerman1964, koch1982, hopfield1984, fromherz1993, hush1999, sima2002}, the capacity of which recently drives the renewal interest in neural networks \cite{frei2020, gidon2020, jones2020, yehudai2020}.
To investigate how a small increase in the network complexity, without overparameterization, may impact the proposed instrument, we also complemented the model with a hidden layer $L=1$ of the ReLU neurons $d_l = 5$, where the LIGHT function is applied only to the output. 

%\begin{figure}[h!]
%	\centering
%	%\includegraphics[width=\textwidth]{light_definition_config}
%	\subfloat[One neuron architecture\label{fig:l0}]{%
%		\includegraphics[width=0.58\columnwidth]{light0}}
%	\hfill
%	\subfloat[One layer architecture\label{fig:l1}]{%
%			\includegraphics[width=0.77\columnwidth]{light1}}
%		\\
%	\caption{Simple network architectures with the LIGHT function
%	}
%	\label{fig:light_archit} 
%\end{figure}

%\subsection{Optimizers}
We compared four non-adaptive methods - SGD with the -default- configuration (\textbf{sigmoid-sgd}) and SGD with the -r-, -E-, -Er- configurations (\textbf{light-v-sgd}  and \textbf{light-g-sgd})  -  to two popular adaptive methods with the -default- configuration - Adam (\textbf{sigmoid-adam}) and AdaGrad (\textbf{sigmoid-adagrad}). 
For all the optimizers, we used the default parameters, batch size $|B(t)| = 75$, and $n_\mathrm{epoch} = 1500$. The number of runs were equal to 10.

The light-v and light-g hyperparameters were optimized with a random search~\cite{bergstra2012} with a 2.5\% random pick of all possible combinations from the full grid space within the following ranges: $r\in[0.1, 20]$ with the number of points $n_r = 5$; $E\in[0, 20]$, $n_E = 5$; $T\in[0,3]$, $n_T = 3$; $N_T\in[0.2, 0.8]$, $n_{N_T} = 5$. The number of epochs for hyperparameter search was equal to 1. 
As we optimized the hyperparameters with a small percentage of random combinations and one epoch, we added redundancy to the experimental setup by implementing both variations of the light funciton (light-v and light-g). This allowed us to validate the consistency of the chosen hyperparameters and to distinguish the examples where the rates $r$ and $E$ were  properly balanced.    

The LIGHT function in two variations was implemented as a custom output activation layer with Keras \texttt{class LIGHT(Layer)}. The layer controls a convergence and generalization trade-off with the LIGHT configurations. 
The code is available at the repository: \url{https://github.com/yukinoi/light-diagnostic-function}.

\subsection{Synthetic Data}
\begin{figure}[h!]
	\centering
	%\fbox{\rule[-.5cm]{0cm}{4cm} \rule[-.5cm]{4cm}{0cm}}
	\includegraphics[width=1\columnwidth]{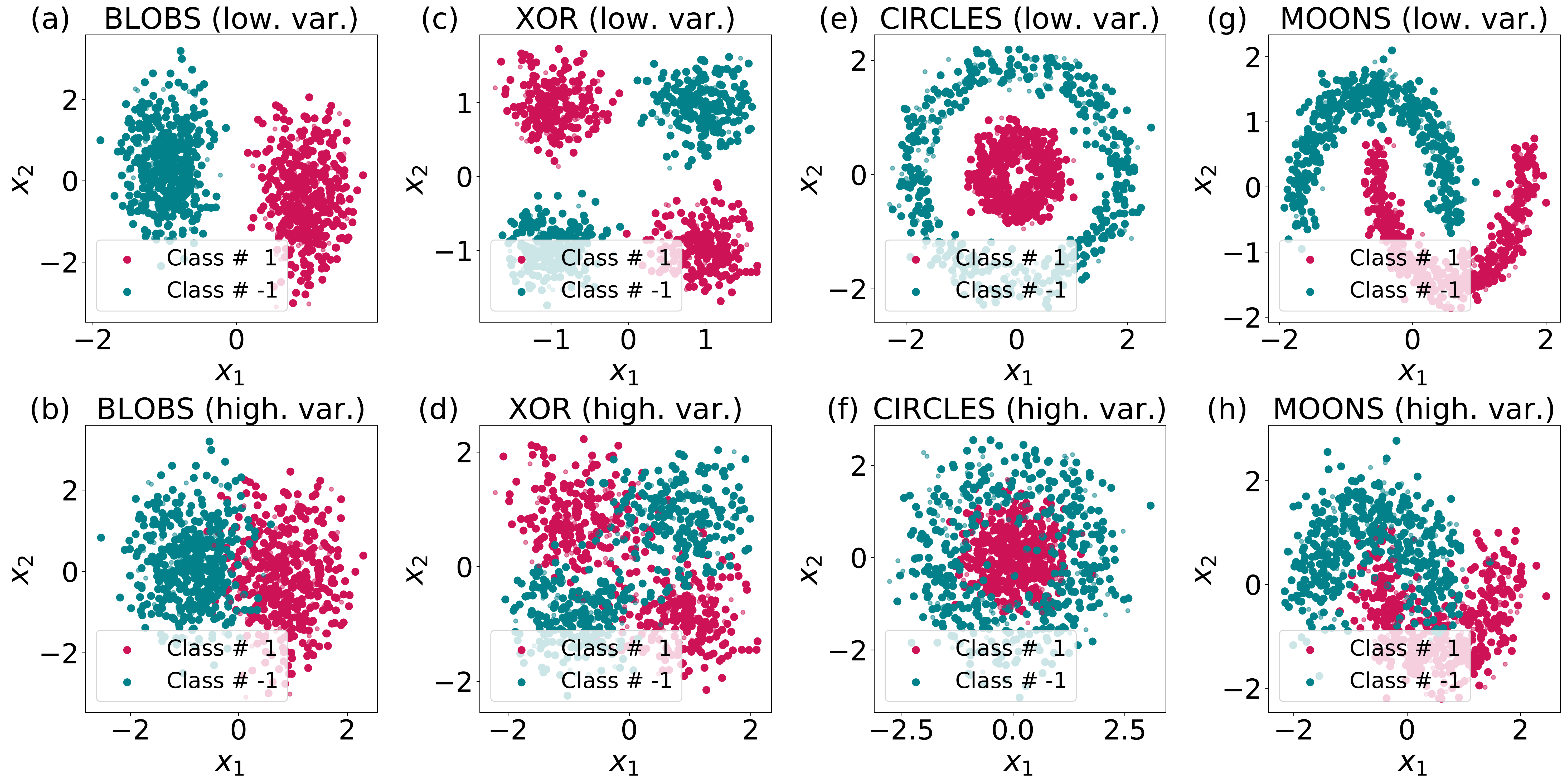}
	\caption{Synthetic datasets.
		(a) BLOBS: \emph{cluster std} = 0.25 (lower variance), (b) BLOBS: \emph{cluster std} = 0.5 (higher variance).
		(c) XOR: \emph{cluster std} = 0.45 (lower variance), (d) XOR: \emph{cluster std} = 0.9 (higher variance).
		(e) CIRCLES: \emph{noise} = 0.1 (lower variance), (f) CIRCLES: \emph{noise} = 0.25 (higher variance).
		(g) MOONS: \emph{noise} = 0.1 (lower variance), (h) MOONS: \emph{noise} = 0.25 (higher variance).}
	\label{fig::syn_dataset}
\end{figure}

We generated a set of synthetic linearly separable and non-separable datasets ($m = 1000$, $n = 2$) with lower and higher levels of variance (see Fig. \ref{fig::syn_dataset}). The datasets were randomly split into training (80\%) and testing (20\%) subsets.

%\textcolor{blue}{\emph{factor} = 0.3}

For conciseness, all the plots on synthetic datasets for different combinations of network architectures are deferred to Appendix~\ref{add::exps}. Figures~\ref{fig::linearL0std25}-\ref{fig::moonsL1std25} (a), (c), (e) depict the test accuracy curves for the -r-, -E-, -Er- configurations (light-v and light-g) on the synthetic datasets with lower and higher variance. For the sake of comparison, we added the -default- configuration of non-adaptive (sigmoid-sgd) and adaptive (sigmoid-adam and sigmoid-adagrad) optimizers to each plot. The presented results comply with the expected behavior of curves given in Figure~\ref{fig:config}. The curves on CIRCLES with one neuron  \ref{fig::circlesL0std25}-\ref{fig::circlesL0std25} seem different as they reach the maximum accuracy in a few epochs in comparison with the other datasets. However, they behave similarly within these epochs. 

Tables~\ref{tbl::blobs}-\ref{tbl::moons} summarize the maximum test accuracy and the number of epochs needed to reach it for each dataset. The best balance between the maximum accuracy and the number of epochs is highlighed in bold. We can observe that both light-v-sgd and light-g-sgd outperform other optimizers with the -default- configuration. The differences between the light variations point to the apparent discrepancy in the balance between $r$ and $E$ occured due to the setting of hyperparameter optimization. When $r$ and $E$ are not fully balanced, we can also observe that the -r- or -E- configuration become slightly superior to the -Er- configuration (see Table \ref{tbl::blobs}, light-g-sgd; Table \ref{tbl::xor}, light-v-sgd; Table \ref{tbl::circles}, light-g-sgd).  

\begin{table*}[h!]
	\caption{Maximum Accuracy And Number of Epochs on BLOBS Dataset}
	\label{tbl::blobs}
	\centering
	%\begin{threeparttable}[b]
	\begin{tabular}{l c l l l l l l l l}
		\toprule
		&  & \multicolumn{4}{c}{L = 0} & \multicolumn{4}{c}{L = 1} \\ 
		\cmidrule{3-6} \cmidrule{7-10} 
		&  & \multicolumn{2}{c}{Lower variance} & \multicolumn{2}{c}{Higher variance} & \multicolumn{2}{c}{Lower variance} & \multicolumn{2}{c}{Higher variance} \\ 
		\cmidrule{3-10} 
		Method & Configuration & Accuracy, \% & $n_{\rm epoch}$ & Accuracy, \% & $n_{\rm epoch}$ & Accuracy, \% & $n_{\rm epoch}$ & Accuracy, \% & $n_{\rm epoch}$\\
		\midrule
		sigmoid-adam& -default- & 100 &349&
		97.45&531 & 
		100 &88&
		97.2&113\\
		%\addlinespace
		sigmoid-adagrad& -default- &99.7&1480&
		96.25&1488 &
		100&196&
		97.2&472\\
		%\addlinespace
		sigmoid-sgd&  -default-  &100&112&
		97.5&330 &
		100&69&
		97.4&232\\
		%\addlinespace
		light-v-sgd&-r-&95.65&1469&
		94.55&1478 &
		100&726&
		96.8&1206\\
		&-E-&100&127&
		97.5&311 &
		100&150&
		97.4&262\\
		&-Er- &\textbf{100}&\textbf{11}&
		\textbf{97.5}&\textbf{47} &
		\textbf{100}&\textbf{67}&
		96.8&818\\
		%\addlinespace
		light-g-sgd&-r-&99.9&769&
		96.7&1139 &
		100&760&
		96.95&1206\\
		&-E-&100&56&
		97.5&137 &
		100&69&
		\textbf{97.45}&\textbf{78}\\
		&-Er-&94.85&7&
		96.8&64 &
		100&94&
		96.65&221\\
		\bottomrule
	\end{tabular}
	%\begin{tablenotes}
	%\item[*] bold font highlights the best balance between the maximum accuracy and the number of epochs
	%\end{tablenotes}
	%\end{threeparttable}
\end{table*}

\begin{table*}[h!]
	\caption{Maximum Accuracy And Number of Epochs on XOR Dataset}
	\label{tbl::xor}
	\centering
	%\begin{threeparttable}[b]
	\begin{tabular}{l c l l l l l l l l}
		\toprule
		&  & \multicolumn{4}{c}{L = 0} & \multicolumn{4}{c}{L = 1} \\ 
		\cmidrule{3-6} \cmidrule{7-10} 
		&  & \multicolumn{2}{c}{Lower variance} & \multicolumn{2}{c}{Higher variance} & \multicolumn{2}{c}{Lower variance} & \multicolumn{2}{c}{Higher variance} \\ 
		\cmidrule{3-10} 
		Method & Configuration & Accuracy, \% & $n_{\rm epoch}$ & Accuracy, \% & $n_{\rm epoch}$ & Accuracy, \% & $n_{\rm epoch}$ & Accuracy, \% & $n_{\rm epoch}$\\
		\midrule
		sigmoid-adam& -default- & 60.5 &401&
		55.2&240& 
		99.95 &1420&
		88.3&450\\
		%\addlinespace
		sigmoid-adagrad& -default- &59.1&1493&
		54.85&1358&
		97.15&1280&
		91.75&1486\\
		%\addlinespace
		sigmoid-sgd&  -default-  &60.5&283&
		55.4&198&
		99.6&296&
		90.05&1163\\
		%\addlinespace
		light-v-sgd&-r-&62.6&18&
		57.6&14&
		99.95&1420
		&\textbf{92.4}&\textbf{425}\\
		&-E-&62.3&162&
		56.4&159&
		\textbf{99.95}&\textbf{554}&
		89.3&1256\\
		&-Er- &68.25&8&
		57.85&14&
		99.7&58&
		92.1&462\\
		%\addlinespace
		light-g-sgd&-r-&68.7&6&
		58.35&10&
		94.55&1489&
		87.7&1106\\
		&-E-&59.75&126&
		56.25&151&
		97.05&171&
		90.3&1417\\
		&-Er-&\textbf{70.75}&\textbf{6}&
		\textbf{60.85}&\textbf{11}&
		89.15&147&
		91.55&580\\
		\bottomrule
	\end{tabular}
	%\begin{tablenotes}
	%\item[*] bold font highlights the best balance between the maximum accuracy and the number of epochs
	%\end{tablenotes}
	%\end{threeparttable}
\end{table*}

\begin{table*}[h!]
	\caption{Maximum Accuracy And Number of Epochs on CIRCLES Dataset}
	\label{tbl::circles}
	\centering
	%\begin{threeparttable}[b]
	\begin{tabular}{l c l l l l l l l l}
		\toprule
		&  & \multicolumn{4}{c}{L = 0} & \multicolumn{4}{c}{L = 1} \\ 
		\cmidrule{3-6} \cmidrule{7-10} 
		&  & \multicolumn{2}{c}{Lower variance} & \multicolumn{2}{c}{Higher variance} & \multicolumn{2}{c}{Lower variance} & \multicolumn{2}{c}{Higher variance} \\ 
		\cmidrule{3-10} 
		Method & Configuration & Accuracy, \% & $n_{\rm epoch}$ & Accuracy, \% & $n_{\rm epoch}$ & Accuracy, \% & $n_{\rm epoch}$ & Accuracy, \% & $n_{\rm epoch}$\\
		\midrule
		sigmoid-adam& -default- & 51.45 &119&
		52.6&95&
		92.75&625&
		87.15&443\\
		%\addlinespace
		sigmoid-adagrad& -default- &52.44&0&
		51.3&710&
		99.65&1423&
		85.7&1143\\
		%\addlinespace
		sigmoid-sgd&  -default-  &49.35&44&
		51.5&237&
		99.95&627&
		87.55&743\\
		%\addlinespace
		light-v-sgd&-r-&49.8&2&
		52.4&3&
		93.65&1023&
		77.55&1499\\
		&-E-&46.65&245&
		51.55&251&
		100&1232&
		87&1177\\
		&-Er- &54.1&1&
		50.7&0&
		99.95&388&
		85.6&88\\
		%\addlinespace
		light-g-sgd&-r-&60.7&1&
		\textbf{56.1}&\textbf{0}&
		100&102&
		82.65&514\\
		&-E-&55.05&34&
		54.75&72&
		96.25&1259&
		86.75&1287\\
		&-Er-&\textbf{61.55}&\textbf{2}&
		55.25&0&
		\textbf{100}&\textbf{69}&
		\textbf{87.56}&\textbf{34}\\
		\bottomrule
	\end{tabular}
	%\begin{tablenotes}
	%\item[*] bold font highlights the best balance between the maximum accuracy and the number of epochs 
	%\end{tablenotes}
	%\end{threeparttable}
\end{table*}

\begin{table*}[h!]
	\caption{Maximum Accuracy And Number of Epochs on MOONS Dataset}
	\label{tbl::moons}
	\centering
	%\begin{threeparttable}[b]
	\begin{tabular}{l c l l l l l l l l}
		\toprule
		&  & \multicolumn{4}{c}{L = 0} & \multicolumn{4}{c}{L = 1} \\ 
		\cmidrule{3-6} \cmidrule{7-10} 
		&  & \multicolumn{2}{c}{Lower variance} & \multicolumn{2}{c}{Higher variance} & \multicolumn{2}{c}{Lower variance} & \multicolumn{2}{c}{Higher variance} \\ 
		\cmidrule{3-10} 
		Method & Configuration & Accuracy, \% & $n_{\rm epoch}$ & Accuracy, \% & $n_{\rm epoch}$ & Accuracy, \% & $n_{\rm epoch}$ & Accuracy, \% & $n_{\rm epoch}$\\
		\midrule
		sigmoid-adam& -default- & 
		88.5 &639&
		85.55&617&
		97.2&849&
		92.75&1275\\
		%\addlinespace
		sigmoid-adagrad& -default- &
		85.45&1486&
		82.8&1477&
		91.65&1459&
		87.7&1469\\
		%\addlinespace
		sigmoid-sgd&  -default-  &
		88.5&436&
		85.75&428&
		94.1&1491&
		90.9&1496\\
		%\addlinespace
		light-v-sgd
		&-r-&
		84.35&1475&
		84&1461&
		95.95&1407&
		91.75&900\\
		&-E-&
		88.5&458&
		85.75&444&
		93.6&1490&
		90&1493\\
		&-Er- &
		\textbf{88.5}&\textbf{47}&
		\textbf{85.75}&\textbf{36}&
		97.95&270&
		\textbf{94.45}&\textbf{624}\\
		%\addlinespace
		light-g-sgd
		&-r-&
		86.7&1477&
		83.55&1481&
		99.9&1469&
		93.5&704\\
		&-E-&
		88.5&297&
		84&46&
		99.45&1339&
		92.35&1495\\
		&-Er-&
		87.2&31&
		83.65&1455&
		\textbf{98.75}&\textbf{687}&
		93.65&1276\\
		\bottomrule
	\end{tabular}
	%\begin{tablenotes}
	%\item[*] bold font highlights the best balance between the maximum accuracy and the number of epochs
	%\end{tablenotes}
	%\end{threeparttable}
\end{table*}
  
The optimal light-v and light-g hyperparameters for different configurations are demonstrated in Figures~\ref{fig::linearL0std25}-\ref{fig::moonsL1std25} (b), (d), (f). By analyzing the boxplots, we see that a non-zero decline rate in the -Er- configuration increases the growth rate $r$ compared to the value $r$ in the -r- configuration. It also brings more stable results as the standard deviation of accuracy curves is substaintially reduced.

When optimizing $r$ on training (validation), the result does not deliver good generalization. This means that the system is pushed towards the edge of stability, which is not clearly defined. 
By increasing $E$, we fixed the edge of stability, improving generalization. In addition, it allowed us to shift the region for picking $r$ to the right, allowing for more extreme values, and, thus, accelerating convergence.

To underline the LIGHT benefits in trading off convergence and generalization, we also analyzed the number of epochs at a test accuracy threshold (see Tables~\ref{tbl::capacity_blobs}-\ref{tbl::capacity_moons}). The lowest number of epochs needed to reach the accuracy threshold is strengthened with bold font. The hyphen indicates that the accuracy threshold is not reached in 1500 iterations. We can see that the light-based SGD greatly outperfoms other optimization methods. 

\begin{table}[h!]
	\caption{Number of Epochs at Accuracy Threshold on BLOBS Dataset}
	%97\% 
	\label{tbl::capacity_blobs}
	\setlength{\tabcolsep}{3.5pt}
	\centering
	%\begin{threeparttable}[b]
		\begin{tabular}{l c l l l l l l}
			\toprule
			&  & \multicolumn{2}{c}{Acc. $\geq$ 95\%, L = 0} & \multicolumn{2}{c}{Acc. $\geq$ 95\%, L = 1} \\ 
			\cmidrule{3-6} 
			Method & Config.  & low. var. & high. var.  & low. var.  & high. var.  \\ 
			\midrule
			sigmoid-adam& -default- & 160 & 150 & 35 & 62\\
			%\addlinespace
			sigmoid-adagrad& -default- & 372& 875 & 21 & 61 \\
			%\addlinespace
			sigmoid-sgd&  -default-  & 27 & 45 & 21 & 43  \\
			%\addlinespace
			light-v-sgd
			&-Er- &
			\textbf{4}&6&2&\textbf{7}\\
			%\addlinespace
			light-g-sgd 
			&-Er-&
			-- &\textbf{4}&\textbf{1}&8 \\
			\bottomrule
		\end{tabular}
		%\begin{tablenotes}
		%\item[*] bold font indicates the lowest number of epochs needed to reach the accuracy threshold; 
		%``--'' means the accuracy threshold is not reached in 1500 iterations
		%\end{tablenotes}
	%\end{threeparttable}
\end{table}

\begin{table}[h!]
	\caption{Number of Epochs at Accuracy Threshold on XOR Dataset}
	%97\% 
	\label{tbl::capacity_xor}
	\setlength{\tabcolsep}{3.5pt}
	\centering
	%\begin{threeparttable}[b]
		\begin{tabular}{l c l l l l l l}
			\toprule
			&  & \multicolumn{2}{c}{Acc. $\geq$ 60\%, L = 0} & \multicolumn{2}{c}{Acc. $\geq$ 90\%, L = 1} \\ 
			\cmidrule{3-6} 
			Method & Config.  & low. var. & high. var. & low. var. & high. var.\\ 
			\midrule
			sigmoid-adam& -default- & 290 & --& 36& -- \\
			%\addlinespace
			sigmoid-adagrad& -default- & -- & -- & 70 & 368 \\
			%\addlinespace
			sigmoid-sgd&  -default-  & 243 & -- & 63 & 1100 \\
			%\addlinespace
			light-v-sgd
			&-Er- &
			1 &--&\textbf{2}&53 \\
			%\addlinespace
			light-g-sgd 
			&-Er-&
			\textbf{0} &\textbf{4}&--&\textbf{23} \\
			\bottomrule
		\end{tabular}
		%\begin{tablenotes}
		%	\item[*] bold font indicates the lowest number of epochs needed to reach the accuracy threshold; 
		%	``--'' means the accuracy threshold is not reached in 1500 iterations
		%\end{tablenotes}
	%\end{threeparttable}
\end{table}

\begin{table}[h!]
	\caption{Number of Epochs at Accuracy Threshold on CIRCLES Dataset}
	%97\% 
	\label{tbl::capacity_circles}
	\setlength{\tabcolsep}{3.5pt}
	\centering
	%\begin{threeparttable}[b]
		\begin{tabular}{l c l l l l l l}
			\toprule
			&  & \multicolumn{2}{c}{Acc. $\geq$ 55\%, L = 0} & \multicolumn{2}{c}{Acc. $\geq$ 85\%, L = 1} \\ 
			\cmidrule{3-6} 
			Method & Config.  & low. var. & high. var. & low. var. & high. var.\\ 
			\midrule
			sigmoid-adam& -default- & -- & --& 69& 221 \\
			%\addlinespace
			sigmoid-adagrad& -default- & -- & -- & 67 & 466 \\
			%\addlinespace
			sigmoid-sgd&   -default-  & -- & -- & 77 & 315 \\
			%\addlinespace
			light-v-sgd
			&-Er- &
			-- &--&\textbf{2}& 51 \\
			%\addlinespace
			light-g-sgd 
			&-Er-&
			\textbf{0} &\textbf{0}&3&\textbf{13} \\
			\bottomrule
		\end{tabular}
		%\begin{tablenotes}
		%	\item[*] bold font indicates the lowest number of epochs needed to reach the accuracy threshold; 
		%	``--'' means the accuracy threshold is not reached in 1500 iterations
		%\end{tablenotes}
	%\end{threeparttable}
\end{table}

\begin{table}[h!]
	\caption{Number of Epochs at Accuracy Threshold on MOONS Dataset}
	%97\% 
	\label{tbl::capacity_moons}
	\setlength{\tabcolsep}{3.5pt}
	\centering
	%\begin{threeparttable}[b]
		\begin{tabular}{l c l l l l l l}
			\toprule
			&  & \multicolumn{2}{c}{Acc. $\geq$ 85\%, L = 0} & \multicolumn{2}{c}{Acc. $\geq$ 90\%, L = 1} \\ 
			\cmidrule{3-6} 
			Method & Config.  & low. var. & high. var. & low. var. & high. var.\\ 
			\midrule
			sigmoid-adam& -default- & 244 &310& 256& 500 \\
			%\addlinespace
			sigmoid-adagrad& -default- & 1306 & -- & 780 & -- \\
			%\addlinespace
			sigmoid-sgd&  -default-  & 47 & 172 & 580 & 1379 \\
			%\addlinespace
			light-v-sgd
			&-Er- &
			\textbf{2} &\textbf{15}&24& \textbf{29} \\
			%\addlinespace
			light-g-sgd 
			&-Er-&
			5 &--&\textbf{5}&58 \\
			\bottomrule
		\end{tabular}
		%\begin{tablenotes}
		%	\item[*] bold font indicates the lowest number of epochs needed to reach the accuracy threshold; 
		%	``--'' means the accuracy threshold is not reached in 1500 iterations
		%\end{tablenotes}
	%\end{threeparttable}
\end{table}

\subsection{Application}

We validated the proposed step size adaptation approach with the LIGHT diagnostic function on MNIST, Fashion MNIST, and CIFAR10 datasets. The labels of the image classification datasets were binarized with the target class \{5\}. The samples were randomly extracted ($m=1000$) from each of them and split into training (80\%) and testing (20\%) subsets. To classify the images, we used the one-layer network architecture and the light-g variation with the pre-defined growth and decline rate: $r = 4.08$, $E = 6.4$.

%Figures~\ref{fig:exp:1} and \ref{fig:exp:2} show the accuracy curves on the test subset for the -Er- configuration. A rough estimate suggests that the dataset \emph{breast cancer win} with the most balanced combination $r$ and $E$  ($E \leq \frac{r}{2}$ for LIGHT-V and $E \leq  r$ for LIGHT-G) demonstrates the best results. The optimal hyperparameters are also the closest to the pre-defined values $E^*$ and $H^*$. The other datasets mostly require a larger pre-defined value $r$ to compensate for a higher harvesting rate $E$. In Appendix~\ref{exp::exp}, we provided a more detailed empirical study with all quantitative results.

\begin{figure}[h!]
	\centering
	%\fbox{\rule[-.5cm]{0cm}{4cm} \rule[-.5cm]{4cm}{0cm}}
	\includegraphics[width=1\columnwidth]{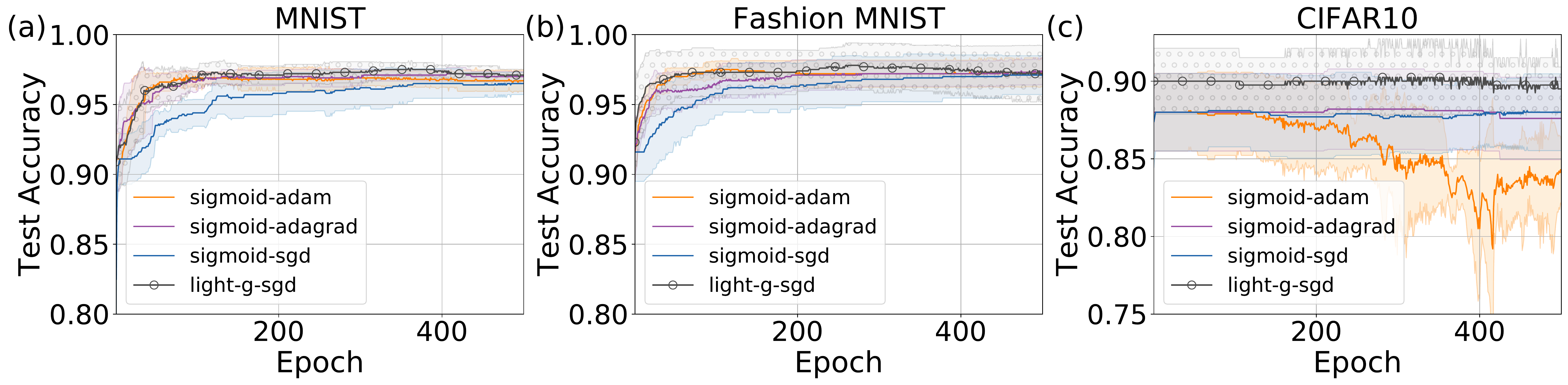}
	\caption{Test accuracy on MNIST, Fashion MNIST, and CIFAR10 datasets for $L = 1$, $d_l = 5$, $m = 1000$}
	\label{fig::imagesL1}
\end{figure}

Figure~\ref{fig::imagesL1} shows the accuracy curves on testing for the -Er- configuration.
The number of epochs needed to reach the maximum accuracy and the test accuracy threshold for each image dataset are summarized in Tables \ref{tbl::mnist}, \ref{tbl::fmnist}, and \ref{tbl::cifar10}. By analogy, bold font indicates the lowest number of epochs needed to reach the accuracy threshold. The hyphen shows that the threshold is not reached.
As we can see, the provided results disclose the benefits of the proposed step size adaptation in managing the trade-off between convergence and generalization with reliable and explainable network architectures.

%\begin{table*}[h!]
%	\caption{Results on Experimental Datasets}
%	%97\% 
%	\label{tbl::mnist_app}
%	%\setlength{\tabcolsep}{1.85pt}
%	\centering
%	%\begin{threeparttable}[b]
%	\begin{tabular}{l l l l l l l l l l}
%		\toprule
%		&  \multicolumn{3}{c}{MNIST} & \multicolumn{3}{c}{Fashion MNIST} & \multicolumn{3}{c}{CIFAR10} \\ 
%		\cmidrule{2-10} 
%		Method & Acc., \% & $n_{\rm epoch} $ & Acc. $\geq$ 97.2\% & Acc., \% & $n_{\rm epoch} $ & Acc. $\geq$  97.6\%& Acc., \% & $n_{\rm epoch} $ & Acc. $\geq$ 90\%\\ 
%		\midrule
%		sigmoid-adam&  97.2 & 1312 & 1312  & 97.5 & 94 & -- & 88.1 & 43 & --\\
%		%\addlinespace
%		sigmoid-adagrad&  97.2 & 405 & 405 & 97.4 & 473 & -- & 88.2 & 213 & --\\
%		%\addlinespace
%		sigmoid-sgd&   97.1 & 952 & -- &  97.6 & 1319 & 1319 & 88.1 & 67 & --\\
%		%\addlinespace
%		light-g-sgd &
%		\bf{97.6} &\bf{358}& \bf{102} & \bf{97.8} &\bf{258} & \bf{235} & \bf{90.5} &  \bf{887} & \bf{0}\\
%		%\addlinespace
%		\bottomrule
%	\end{tabular}
%	%\begin{tablenotes}
%		%\item[*] bold font indicates the lowest number of epochs needed to reach the accuracy threshold; 
%		%``--'' means the accuracy threshold is not reached in 1500 iterations
%	%\end{tablenotes}
%	%\end{threeparttable}
%\end{table*}

\begin{table}[h!]
	\caption{Experimental Results on on MNIST}
	%97\% 
	\label{tbl::mnist}
	\centering
	%\begin{threeparttable}[b]
	\begin{tabular}{l l l l}
		\toprule
		Method & Acc., \% & $n_{\rm epoch} $ & Acc. $\geq$ 97.2\% \\ 
		\midrule
		sigmoid-adam&  97.2 & 1312 & 1312  \\
		%\addlinespace
		sigmoid-adagrad&  97.2 & 405 & 405 \\
		%\addlinespace
		sigmoid-sgd&   97.1 & 952 & -- \\
		%\addlinespace
		light-g-sgd &
		\bf{97.6} &\bf{358}& \bf{102} \\
		%\addlinespace
		\bottomrule
	\end{tabular}
	%\begin{tablenotes}
	%\item[*] bold font indicates the lowest number of epochs needed to reach the accuracy threshold; 
	%``--'' means the accuracy threshold is not reached in 1500 iterations
	%\end{tablenotes}
	%\end{threeparttable}
\end{table}

\begin{table}[h!]
	\caption{Experimental Results on Fashion MNIST}
	%97\% 
	\label{tbl::fmnist}
	\centering
	%\begin{threeparttable}[b]
	\begin{tabular}{l l l l}
		\toprule
		Method & Acc., \% & $n_{\rm epoch} $ & Acc. $\geq$ 97.6\% \\ 
		\midrule
		sigmoid-adam& 97.5 & 94 & --  \\
		%\addlinespace
		sigmoid-adagrad&  97.4 & 473 & --  \\
		%\addlinespace
		sigmoid-sgd&   97.6 & 1319 & 1319 \\
		%\addlinespace
		light-g-sgd &
		\bf{97.8} &\bf{258} & \bf{235} \\
		%\addlinespace
		\bottomrule
	\end{tabular}
	%\begin{tablenotes}
	%\item[*] bold font indicates the lowest number of epochs needed to reach the accuracy threshold; 
	%``--'' means the accuracy threshold is not reached in 1500 iterations
	%\end{tablenotes}
	%\end{threeparttable}
\end{table}

\begin{table}[h!]
	\caption{Experimental Results on CIFAR10}
	%97\% 
	\label{tbl::cifar10}
	\centering
	%\begin{threeparttable}[b]
	\begin{tabular}{l l l l}
		\toprule
		Method & Acc., \% & $n_{\rm epoch} $ & Acc. $\geq$ 90\% \\ 
		\midrule
		sigmoid-adam& 88.1 & 43 & -- \\
		%\addlinespace
		sigmoid-adagrad&  88.2 & 213 & -- \\
		%\addlinespace
		sigmoid-sgd&   88.1 & 67 & -- \\
		%\addlinespace
		light-g-sgd &
		\bf{90.5} &  \bf{887} & \bf{0}\\
		%\addlinespace
		\bottomrule
	\end{tabular}
	%\begin{tablenotes}
	%\item[*] bold font indicates the lowest number of epochs needed to reach the accuracy threshold; 
	%``--'' means the accuracy threshold is not reached in 1500 iterations
	%\end{tablenotes}
	%\end{threeparttable}
	\vspace{-3mm}
\end{table}

% Note that the IEEE does not put floats in the very first column
% - or typically anywhere on the first page for that matter. Also,
% in-text middle ("here") positioning is typically not used, but it
% is allowed and encouraged for Computer Society conferences (but
% not Computer Society journals). Most IEEE journals/conferences use
% top floats exclusively. 
% Note that, LaTeX2e, unlike IEEE journals/conferences, places
% footnotes above bottom floats. This can be corrected via the
% \fnbelowfloat command of the stfloats package.

\section{Conclusion}
We contributed to the direction of SGD based optimization with ``painless'' step size adaptation. This technique allows to increase a step size by some fixed growing rate $r$ which is self-stabilized with some fixed declining rate $E$ in order to ensure best balance between convergence and generalization. It equips the optimizer with a simple instrument for explicit control over convergence/generalization trade-off which is the key to building reliable network architectures.

Rather than suggesting another adaptive and non-monotone activation function, we put forward this instrument as a simple diagnostic function. The function adopts sliding modes in line with some fixed growing $r$ and declining $E$ rates to simulate discontinuities and explicitly regulate their influence on convergence and generalization. 
In addition, the LIGHT function relies on the laws of population dynamics as an original s-shaped monotonic function \cite{wilson1972} but exhibits more complex behavior as noted in \cite{defelice1993}. This means that the proposed ``painless'' step size adaptation may open up new opportunities for building not only reliable but explainable neural network architectures \cite{tjoa2020} with greater capacity.

%\textcolor{blue}{Summary on experiments here ...}

%We introduced the LIGHT function to complement inner processes inside neurons in deep networks with growing and harvesting borrowed from population dynamics. This function allows explicit control of the trade-off between inductive biases and convergence rates with two balanced pre-defined values of \emph{per} capita growth and harvesting rates. The proposed LIGHT function increases both convergence rate and inductive bias without overcomplicating optimization processes and overparametrizing models.    

% or


\appendices

%
%
%% you can choose not to have a title for an appendix
%% if you want by leaving the argument blank
\vspace{0mm}
\appendix[Plots on Synthetic Datasets]
\label{add::exps}
%\section*{Tables}

%\section*{Figures}
%\subsection*{BLOBS datatset}
\setcounter{figure}{0}
\counterwithin{figure}{section}
\setcounter{table}{0}
\counterwithin{table}{section}

\begin{figure}[!htb]
	\centering
	%\fbox{\rule[-.5cm]{0cm}{4cm} \rule[-.5cm]{4cm}{0cm}}
	\includegraphics[width=1\columnwidth]{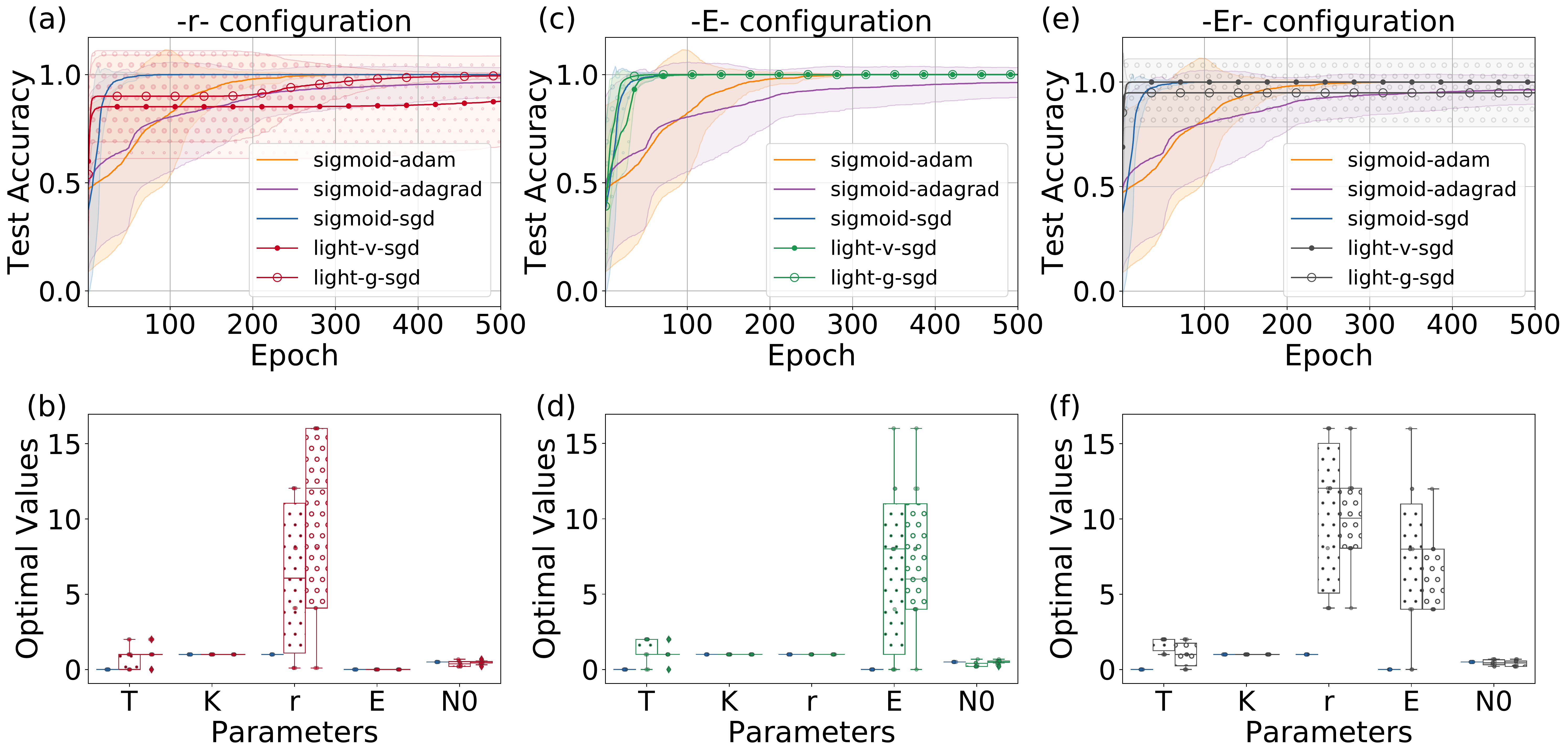}
	\caption{Test accuracy and optimal hyperparameters on BLOBS (lower variance) for $L = 0$, $d_l = 0$, $m = 1000$, $n = 2$, \emph{cluster std} = 0.25}
	\label{fig::linearL0std25}
	\vspace{-2mm}
\end{figure}

\begin{figure}[!htb]
	\centering
	%\fbox{\rule[-.5cm]{0cm}{4cm} \rule[-.5cm]{4cm}{0cm}}
	\includegraphics[width=1\columnwidth]{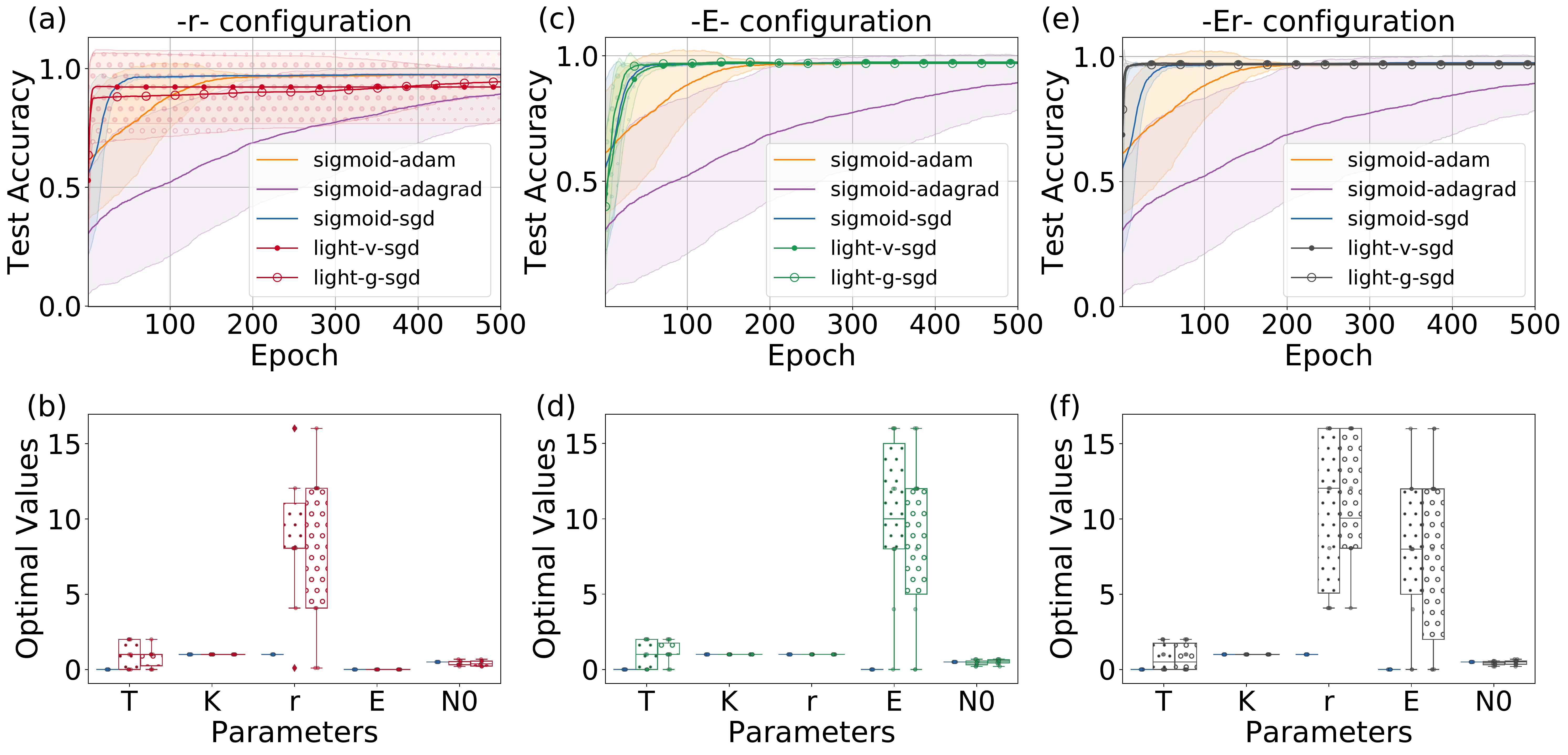}
	\caption{Test accuracy and optimal hyperparameters on BLOBS (higher variance) for $L = 0$, $d_l = 0$, $m = 1000$, $n = 2$, \emph{cluster std} = 0.5}
	\label{fig::linearL0std50}
	\vspace{2mm}
\end{figure}

\begin{figure}[!htb]
	\centering
	%\fbox{\rule[-.5cm]{0cm}{4cm} \rule[-.5cm]{4cm}{0cm}}
	\includegraphics[width=1\columnwidth]{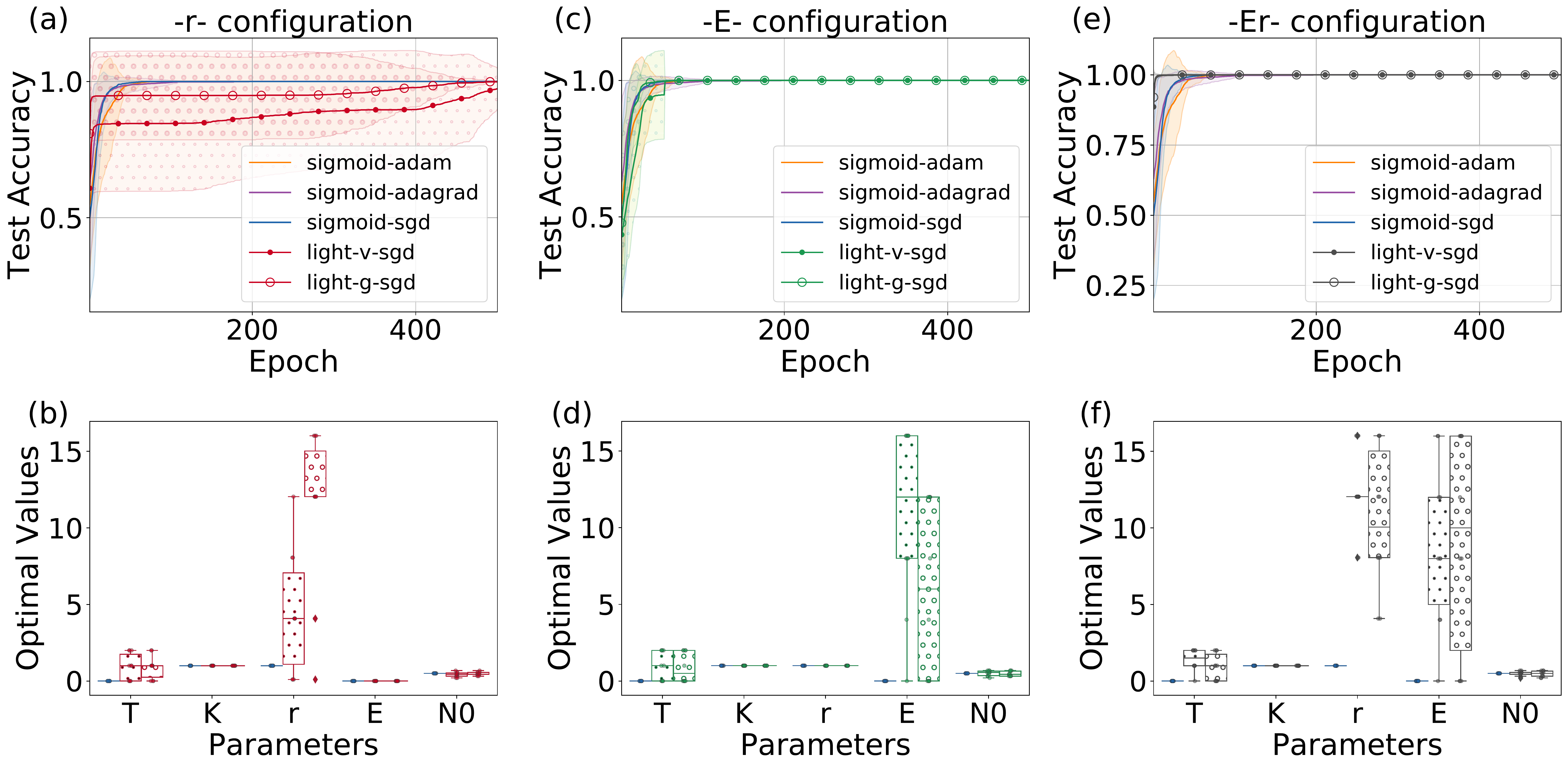}
	\caption{Test accuracy and optimal hyperparameters on BLOBS (lower variance) for $L = 1$, $d_l = 5$, $m = 1000$, $n = 2$, \emph{cluster std} = 0.25}
	\label{fig::linearL1std25}
	\vspace{2mm}
\end{figure}

\begin{figure}[!htb]
	\centering
	%\fbox{\rule[-.5cm]{0cm}{4cm} \rule[-.5cm]{4cm}{0cm}}
	\includegraphics[width=1\columnwidth]{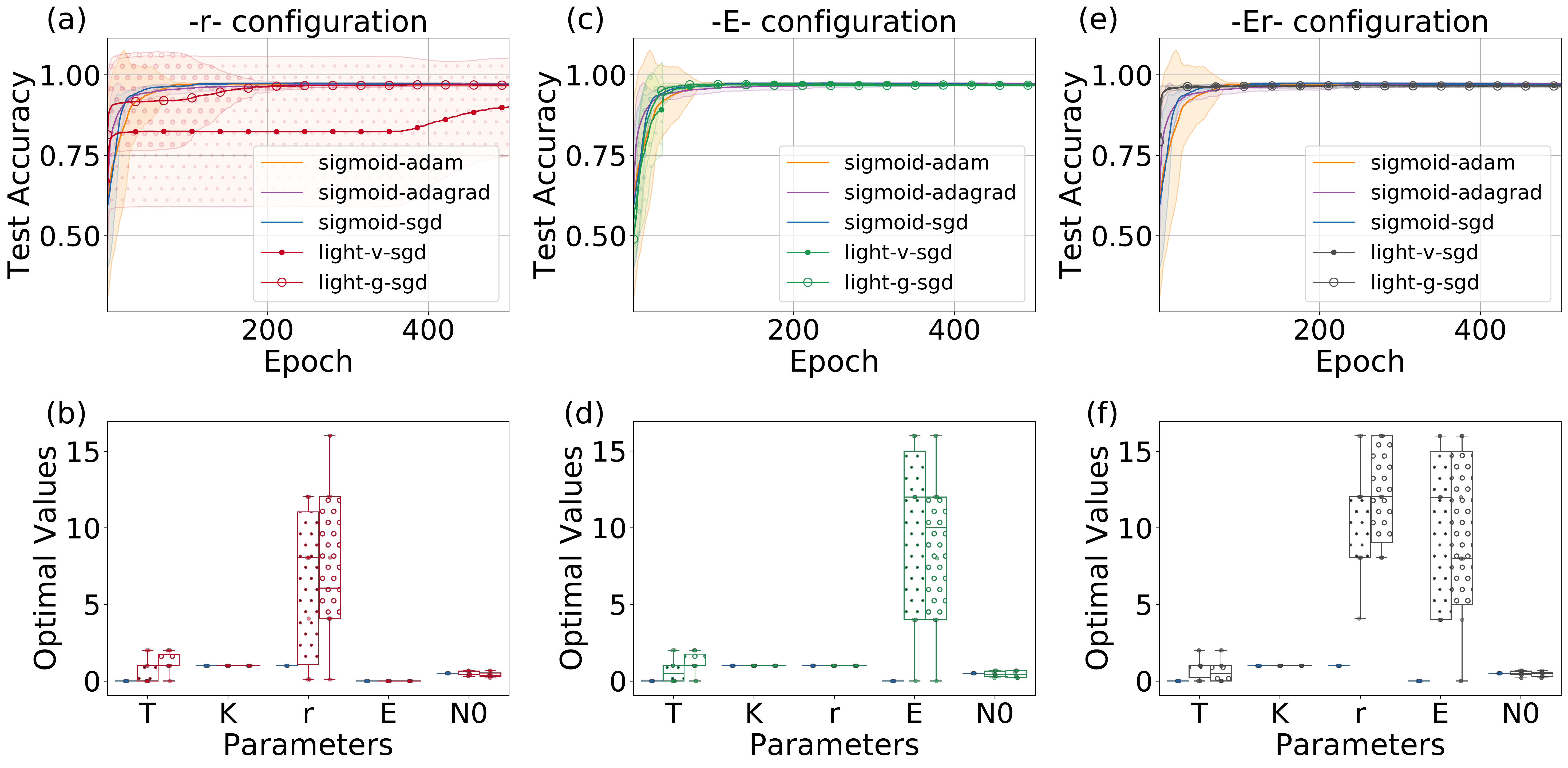}
	\caption{Test accuracy and optimal hyperparameters on BLOBS (higher variance) for $L = 1$, $d_l = 5$, $m = 1000$, $n = 2$, \emph{cluster std} = 0.5}
	\label{fig::linearL1std50}
	\vspace{2mm}
\end{figure}

%\subsection*{XOR datatset}

\begin{figure}[!htb]
	\centering
	%\fbox{\rule[-.5cm]{0cm}{4cm} \rule[-.5cm]{4cm}{0cm}}
	\includegraphics[width=1\columnwidth]{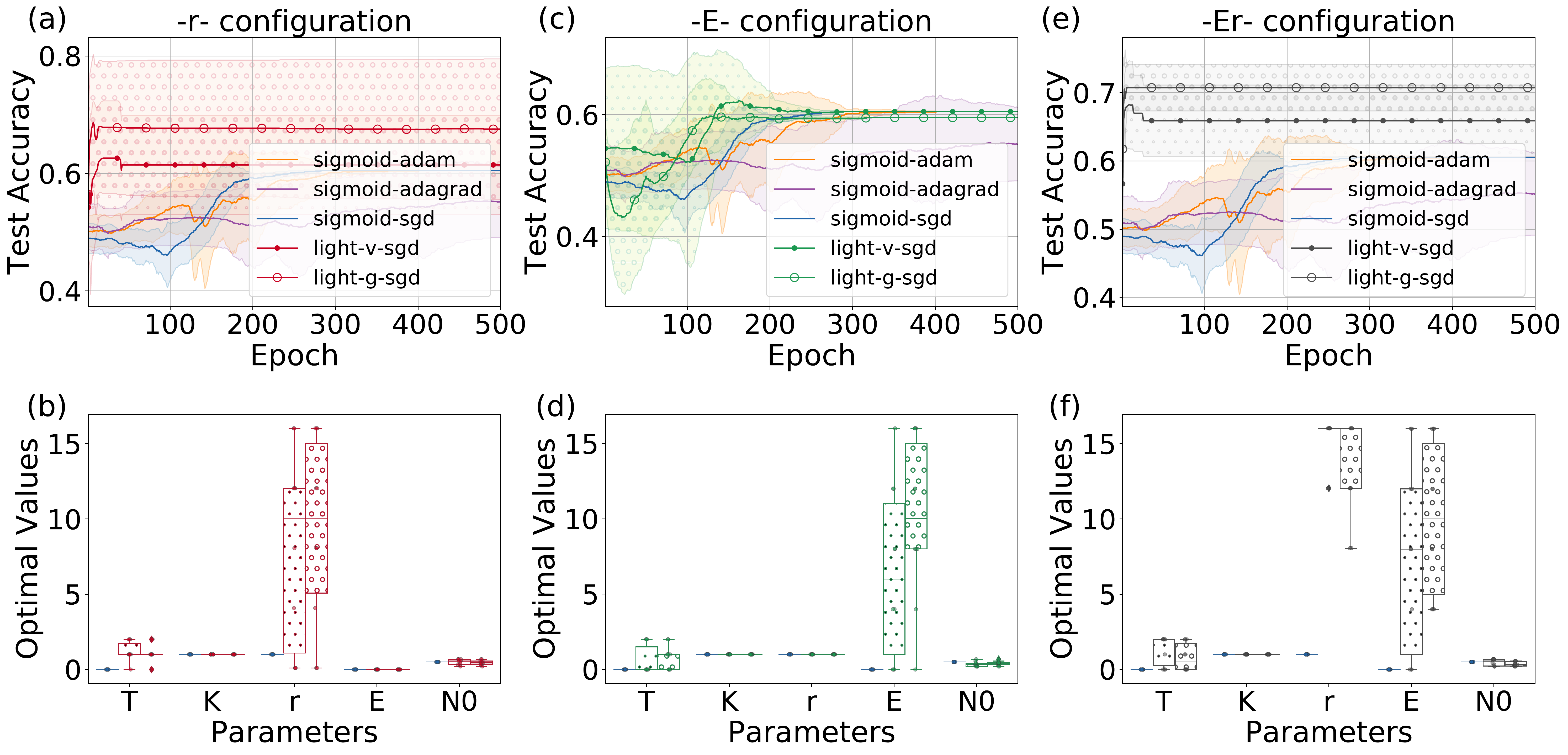}
	\caption{Test accuracy and optimal hyperparameters on XOR (lower variance) for $L = 0$, $d_l = 0$, $m = 1000$, $n = 2$, \emph{cluster std} = 0.45}
	\label{fig::xorL0std45}
	\vspace{2mm}
\end{figure}

\begin{figure}[!htb]
	\centering
	%\fbox{\rule[-.5cm]{0cm}{4cm} \rule[-.5cm]{4cm}{0cm}}
	\includegraphics[width=1\columnwidth]{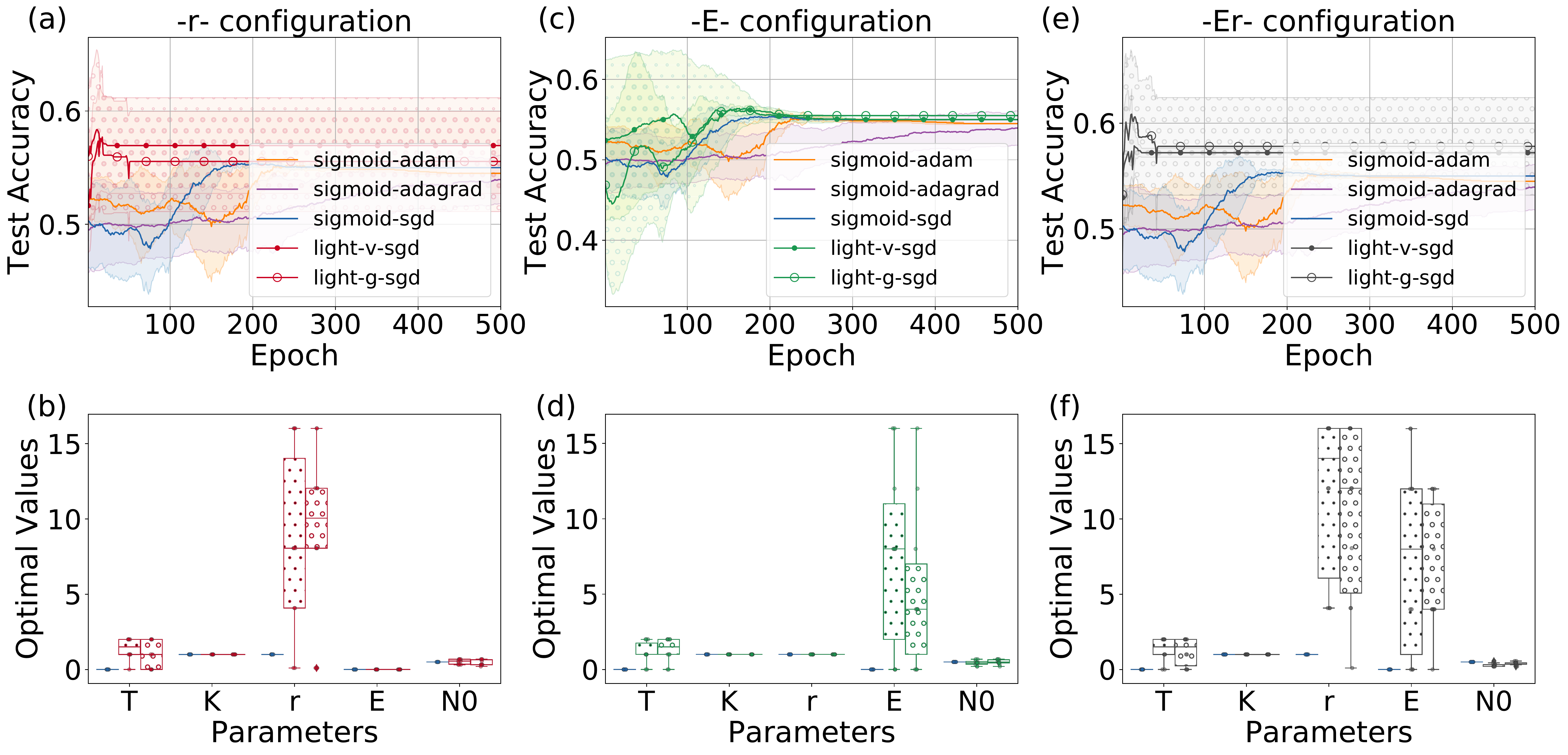}
	\caption{Test accuracy and optimal hyperparameters on XOR (higher variance) for $L = 0$, $d_l = 0$, $m = 1000$, $n = 2$, \emph{cluster std} = 0.90}
	\label{fig::xorL0std90}
	\vspace{2mm}
\end{figure}

\begin{figure}[!htb]
	\centering
	%\fbox{\rule[-.5cm]{0cm}{4cm} \rule[-.5cm]{4cm}{0cm}}
	\includegraphics[width=1\columnwidth]{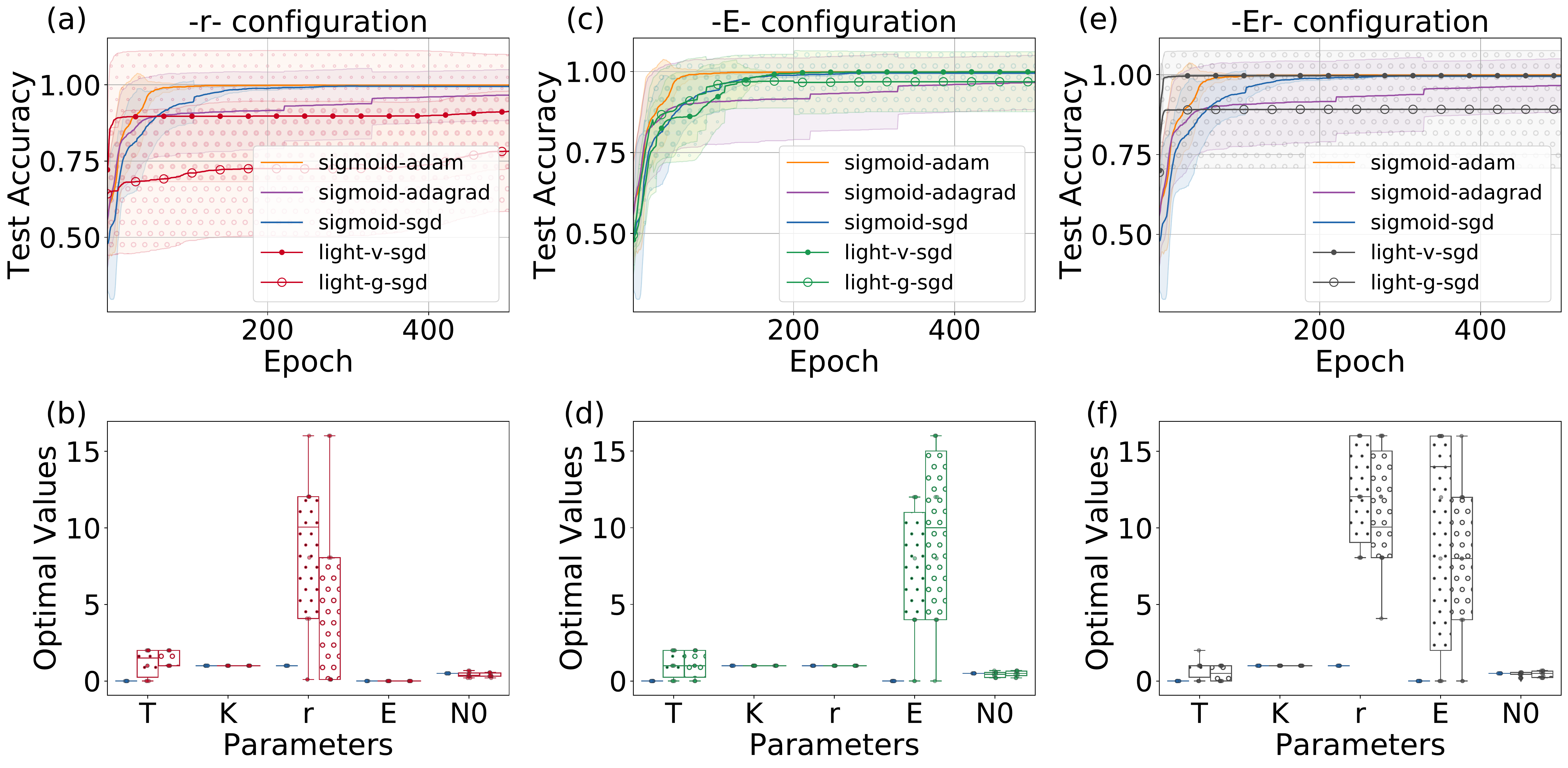}
	\caption{Test accuracy and optimal hyperparameters on XOR (lower variance) for $L = 1$, $d_l = 5$, $m = 1000$, $n = 2$, \emph{cluster std} = 0.45}
	\label{fig::xorL1std45}
	\vspace{2mm}
\end{figure}

\begin{figure}[!htb]
	\centering
	%\fbox{\rule[-.5cm]{0cm}{4cm} \rule[-.5cm]{4cm}{0cm}}
	\includegraphics[width=1\columnwidth]{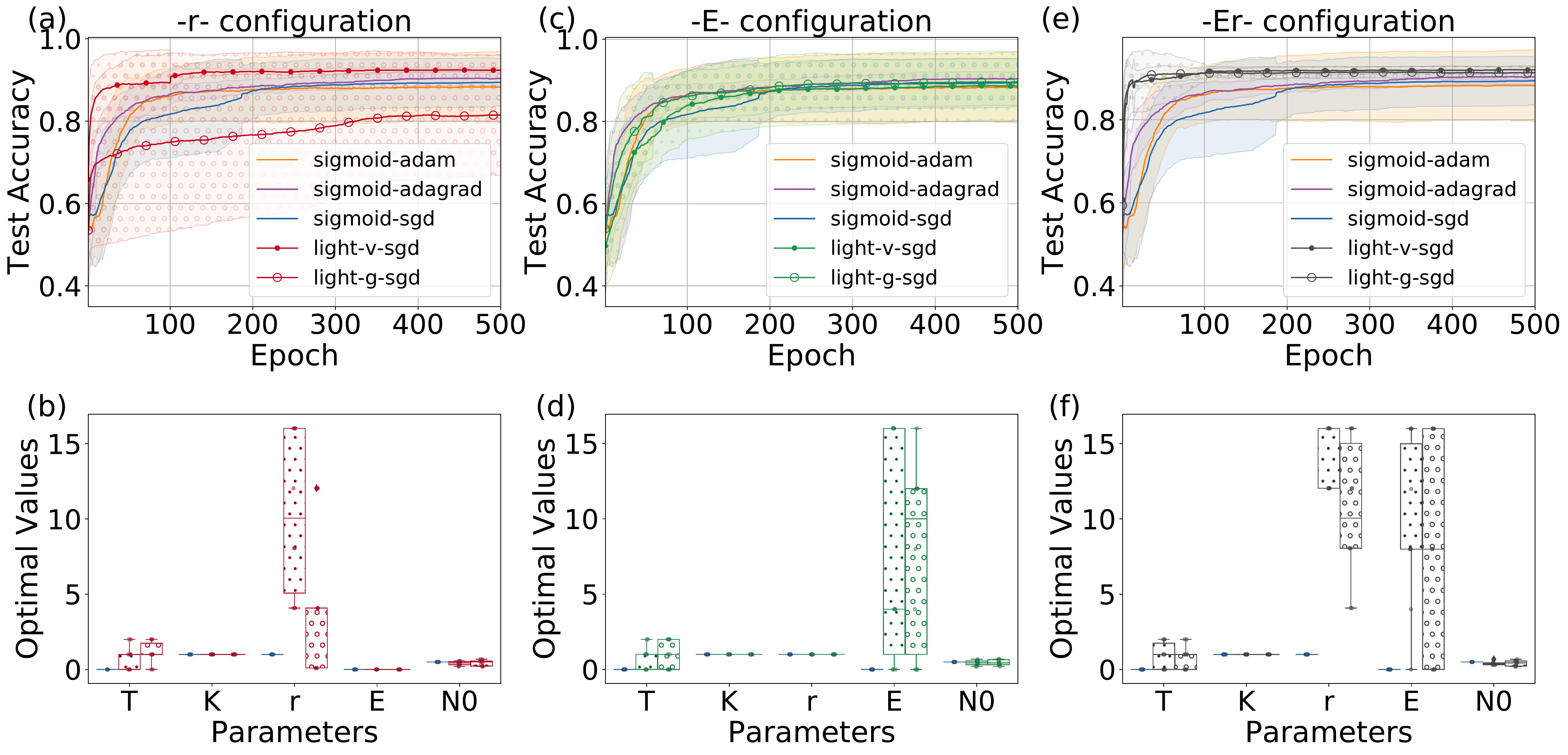}
	\caption{Test accuracy and optimal hyperparameters on XOR (higher variance) for $L = 1$, $d_l = 5$, $m = 1000$, $n = 2$, \emph{cluster std} = 0.9}
	\label{fig::xorL1std90}
	\vspace{2mm}
\end{figure}

%\subsection*{CIRCLES dataset}

\begin{figure}[!htb]
	\centering
	%\fbox{\rule[-.5cm]{0cm}{4cm} \rule[-.5cm]{4cm}{0cm}}
	\includegraphics[width=1\columnwidth]{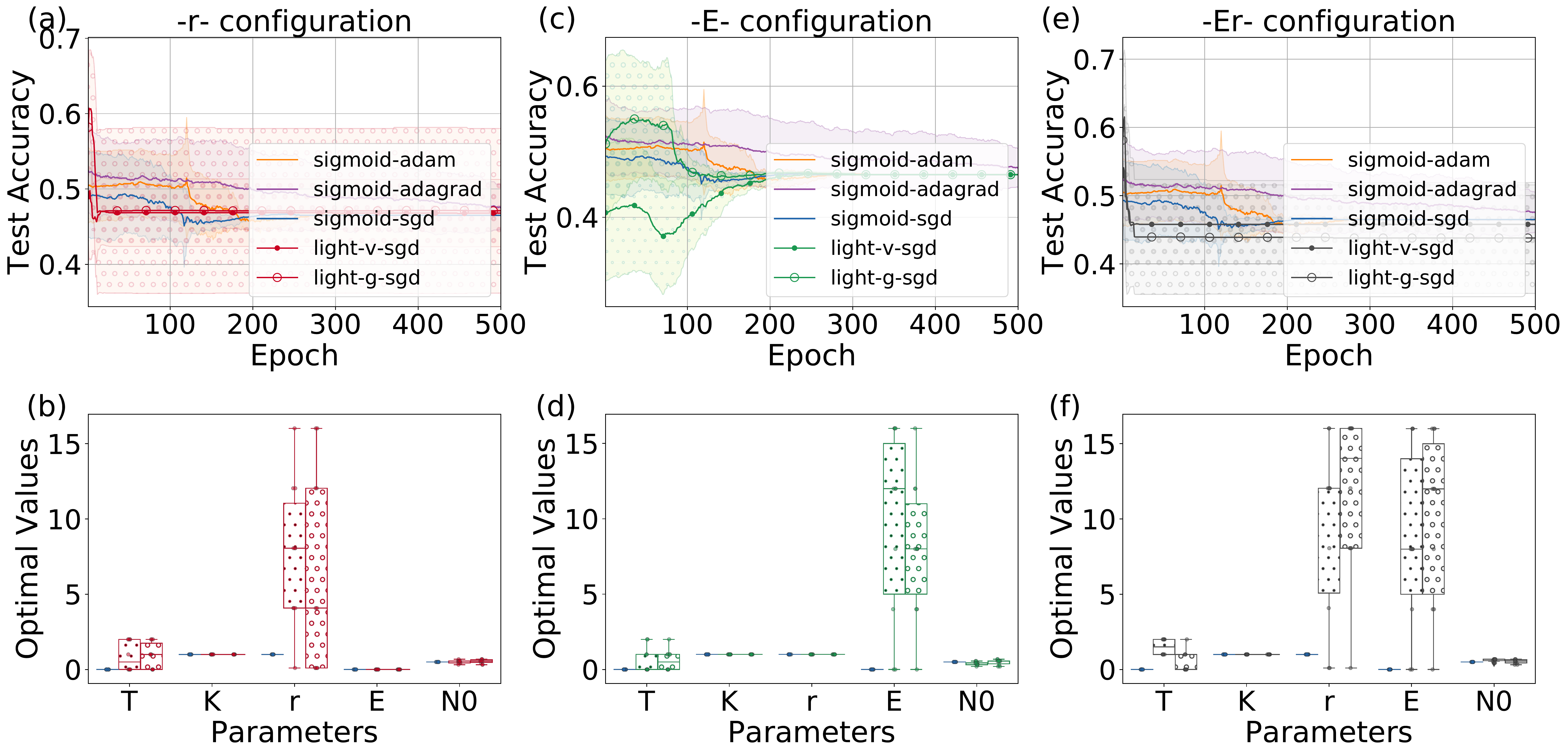}
	\caption{Test accuracy and optimal hyperparameters on CIRCLES (lower variance) for $L = 0$, $d_l = 0$, $m = 1000$, $n = 2$, \emph{noise} = 0.1}
	\label{fig::circlesL0std10}
	\vspace{2mm}
\end{figure}

\begin{figure}[!htb]
	\centering
	%\fbox{\rule[-.5cm]{0cm}{4cm} \rule[-.5cm]{4cm}{0cm}}
	\includegraphics[width=1\columnwidth]{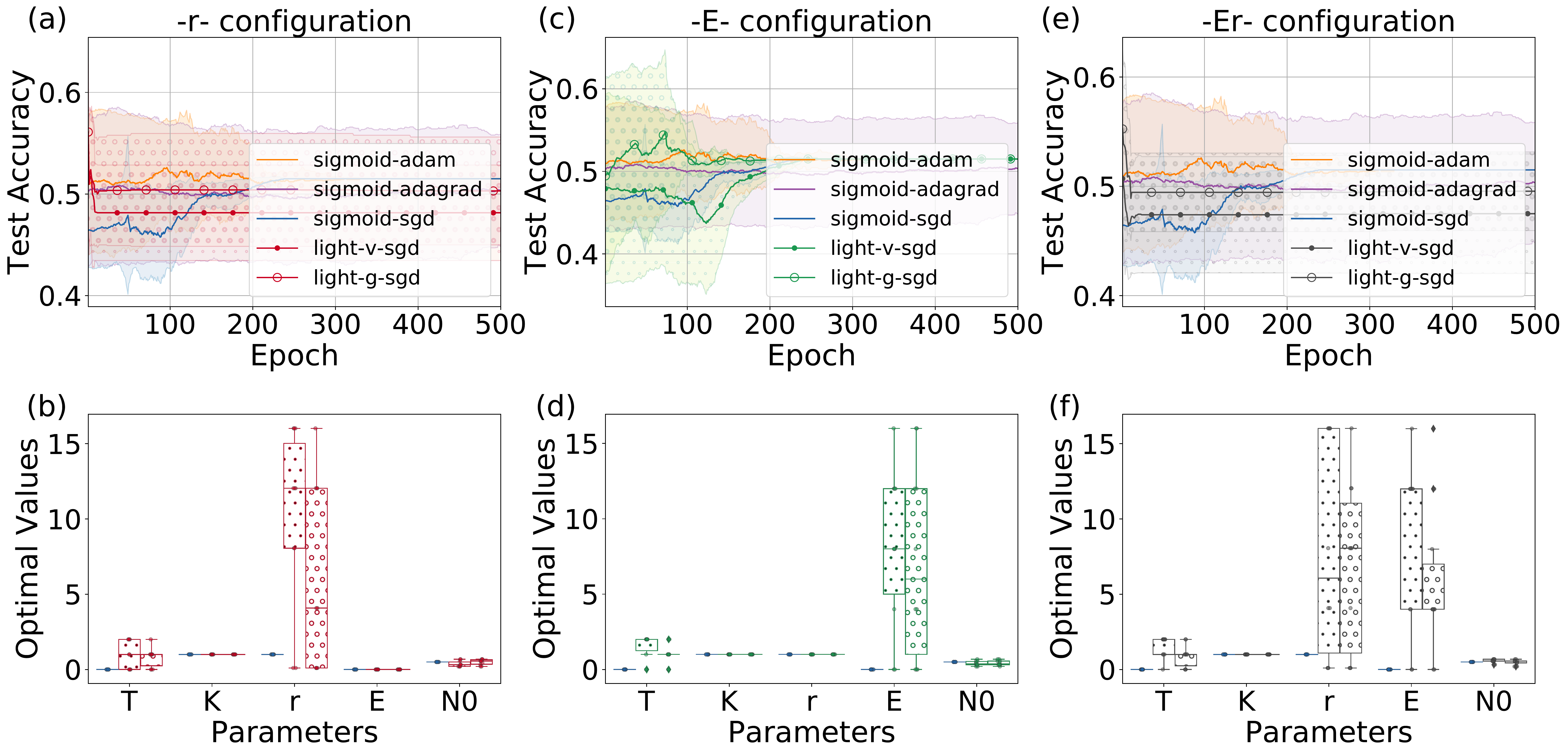}
	\caption{Test accuracy and optimal hyperparameters on CIRCLES (higher variance) for $L = 0$, $d_l = 0$, $m = 1000$, $n = 2$, \emph{noise} = 0.25}
	\label{fig::circlesL0std25}
	\vspace{2mm}
\end{figure}

\begin{figure}[!htb]
	\centering
	%\fbox{\rule[-.5cm]{0cm}{4cm} \rule[-.5cm]{4cm}{0cm}}
	\includegraphics[width=1\columnwidth]{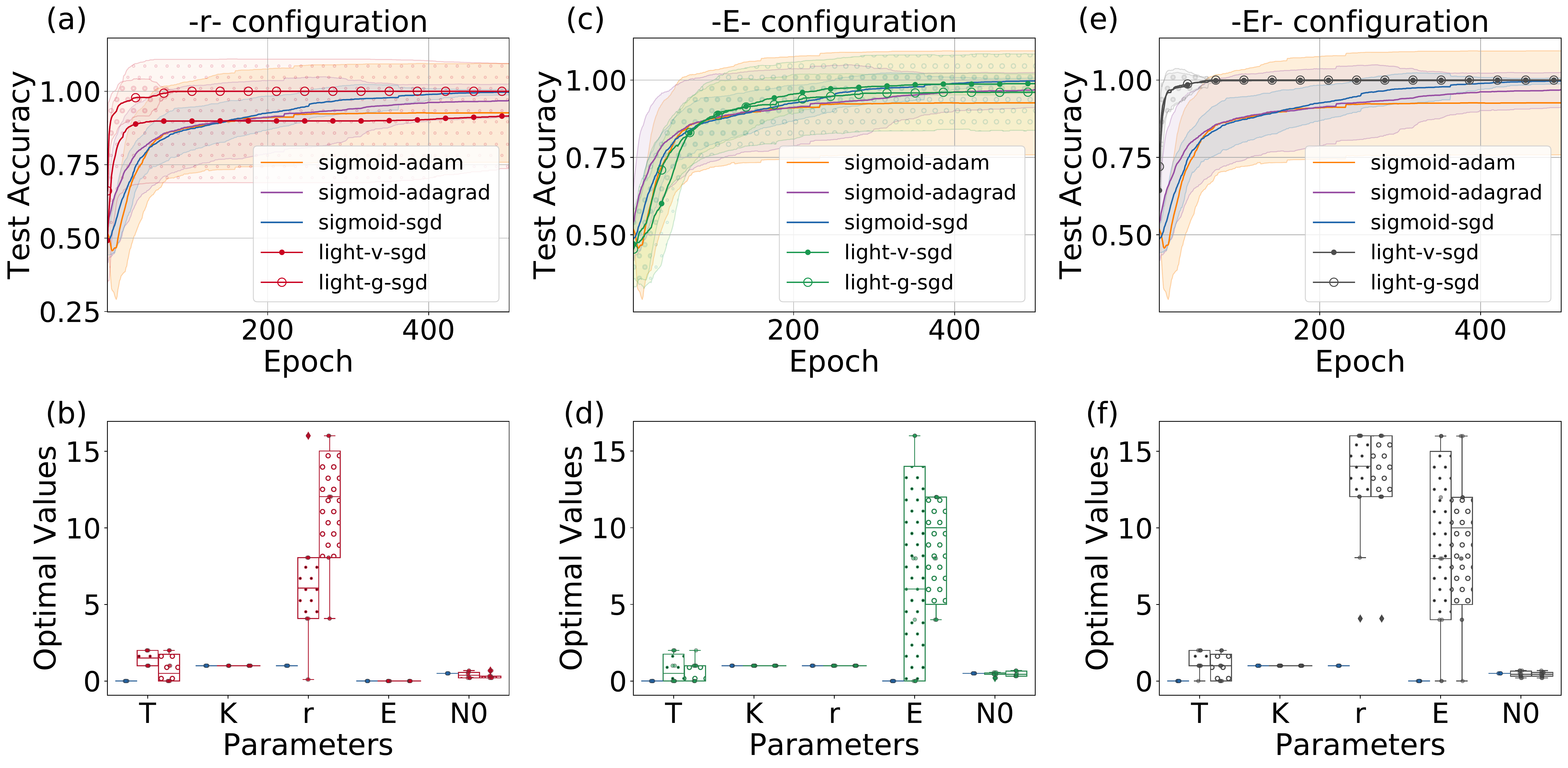}
	\caption{Test accuracy and optimal hyperparameters on CIRCLES (lower variance) for $L = 1$, $d_l = 5$, $m = 1000$, $n = 2$, \emph{noise} = 0.1}
	\label{fig::circlesL1std10}
	\vspace{2mm}
\end{figure}

\begin{figure}[!htb]
	\centering
	%\fbox{\rule[-.5cm]{0cm}{4cm} \rule[-.5cm]{4cm}{0cm}}
	\includegraphics[width=1\columnwidth]{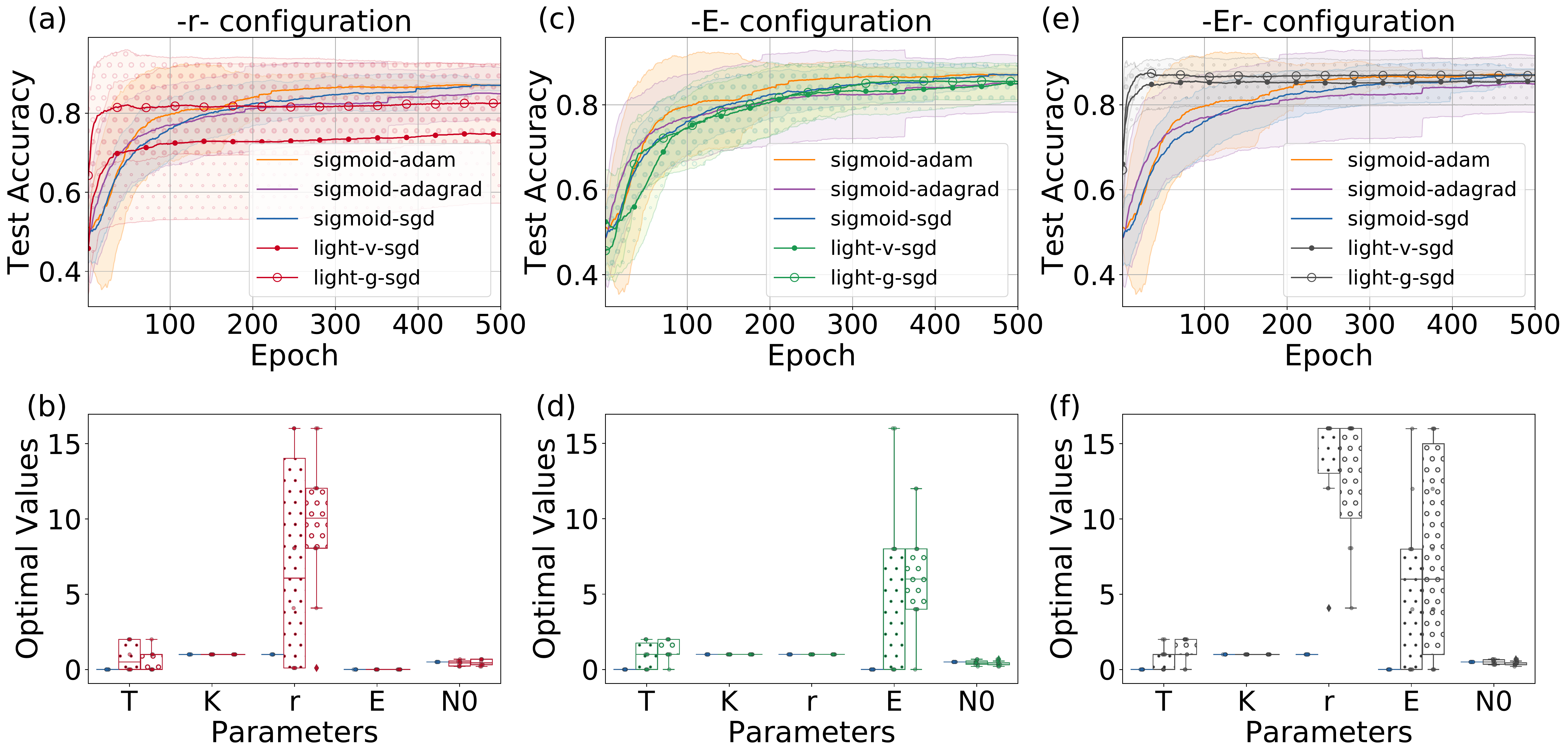}
	\caption{Test accuracy and optimal hyperparameters on CIRCLES (higher variance) for $L = 1$, $d_l = 5$, $m = 1000$, $n = 2$, \emph{noise} = 0.25}
	\label{fig::circlesL1std25}
	\vspace{2mm}
\end{figure}

%\subsection*{MOONS dataset}

\begin{figure}[!htb]
	\centering
	%\fbox{\rule[-.5cm]{0cm}{4cm} \rule[-.5cm]{4cm}{0cm}}
	\includegraphics[width=1\columnwidth]{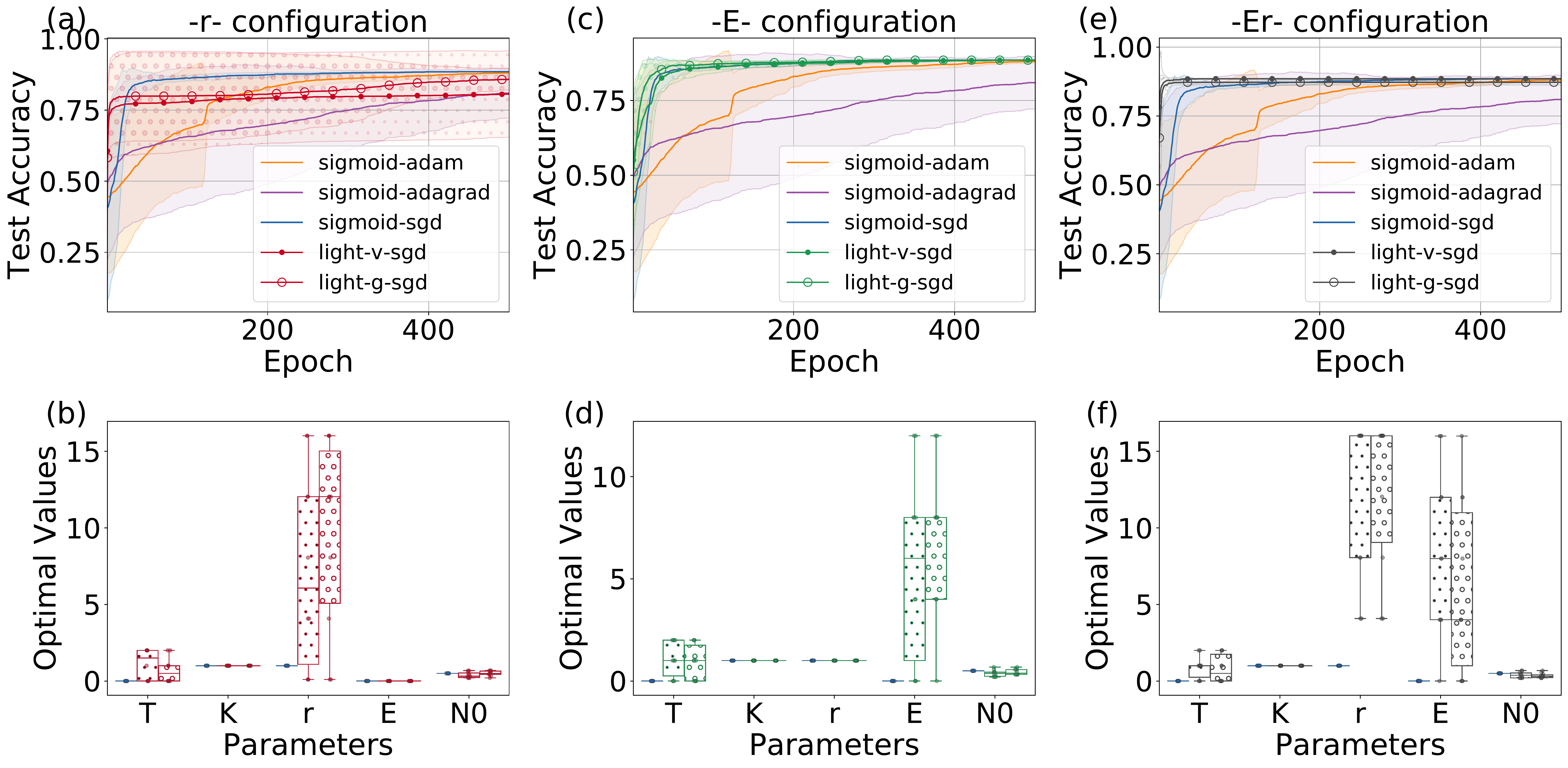}
	\caption{Test accuracy and optimal hyperparameters on MOONS (lower variance) for $L = 0$, $d_l = 0$, $m = 1000$, $n = 2$, \emph{noise} = 0.1}
	\label{fig::moonsL0std10}
	\vspace{2mm}
\end{figure}

\begin{figure}[!htb]
	\centering
	%\fbox{\rule[-.5cm]{0cm}{4cm} \rule[-.5cm]{4cm}{0cm}}
	\includegraphics[width=1\columnwidth]{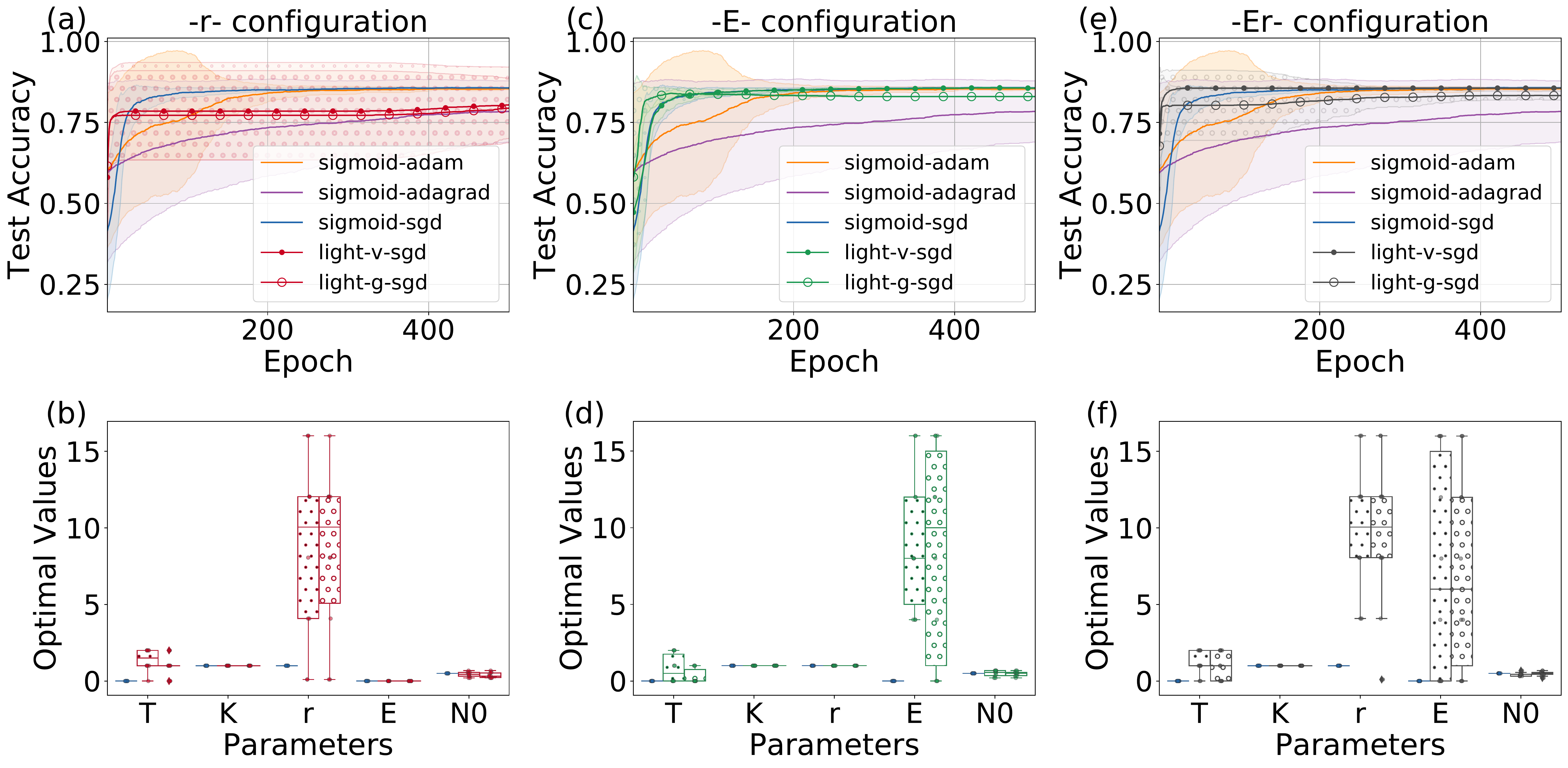}
	\caption{Test accuracy and optimal hyperparameters on MOONS (higher variance) for $L = 0$, $d_l = 0$, $m = 1000$, $n = 2$, \emph{noise} = 0.25}
	\label{fig::moonsL0std25}
	\vspace{2mm}
\end{figure}

\begin{figure}[!htb]
	\centering
	%\fbox{\rule[-.5cm]{0cm}{4cm} \rule[-.5cm]{4cm}{0cm}}
	\includegraphics[width=1\columnwidth]{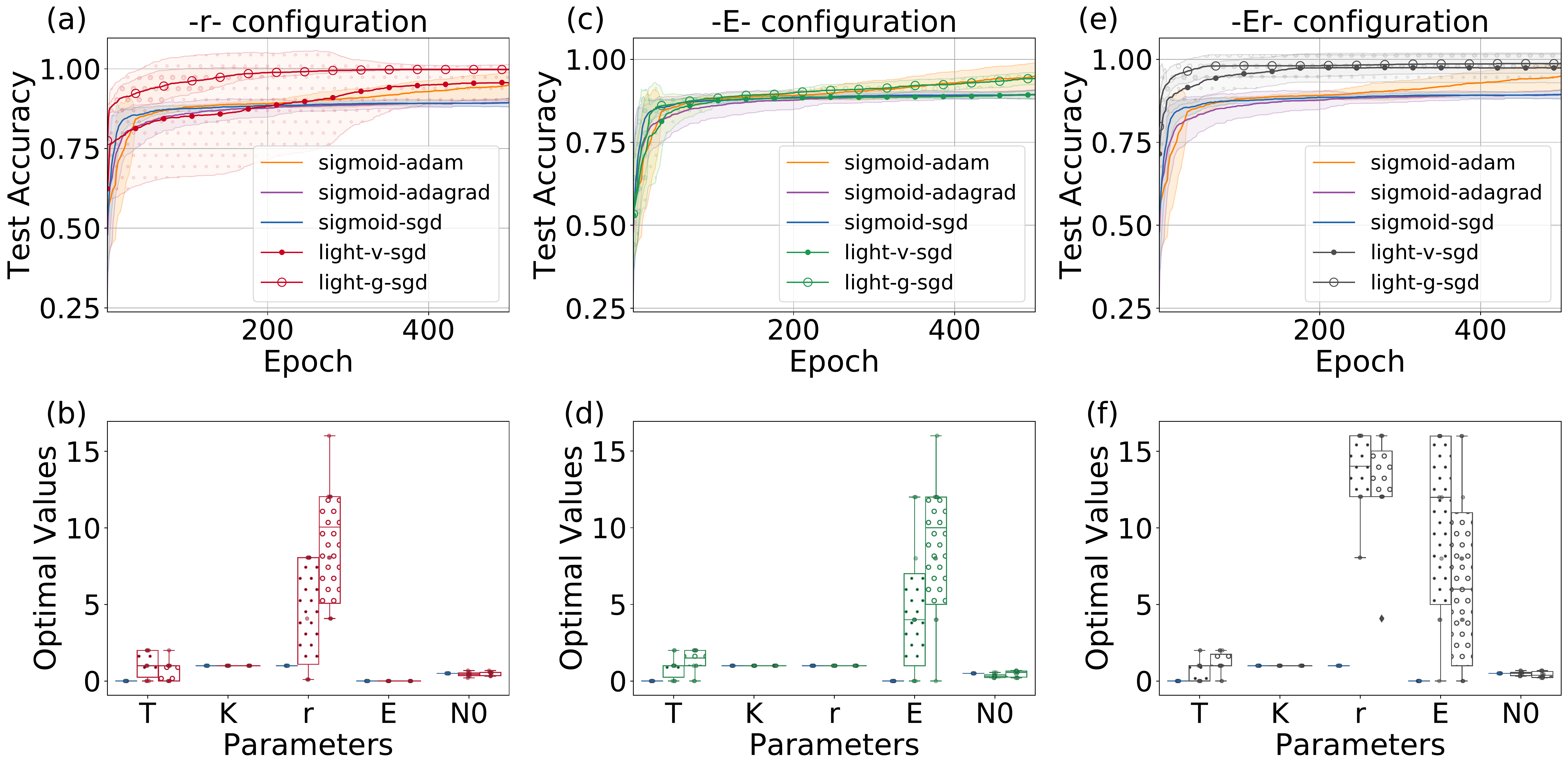}
	\caption{Test accuracy and optimal hyperparameters on MOONS (lower variance) for $L = 1$, $d_l = 5$, $m = 1000$, $n = 2$, \emph{noise} = 0.1}
	\label{fig::moonsL1std10}
	\vspace{2mm}
\end{figure}

\begin{figure}[!htb]
	\centering
	%\fbox{\rule[-.5cm]{0cm}{4cm} \rule[-.5cm]{4cm}{0cm}}
	\includegraphics[width=1\columnwidth]{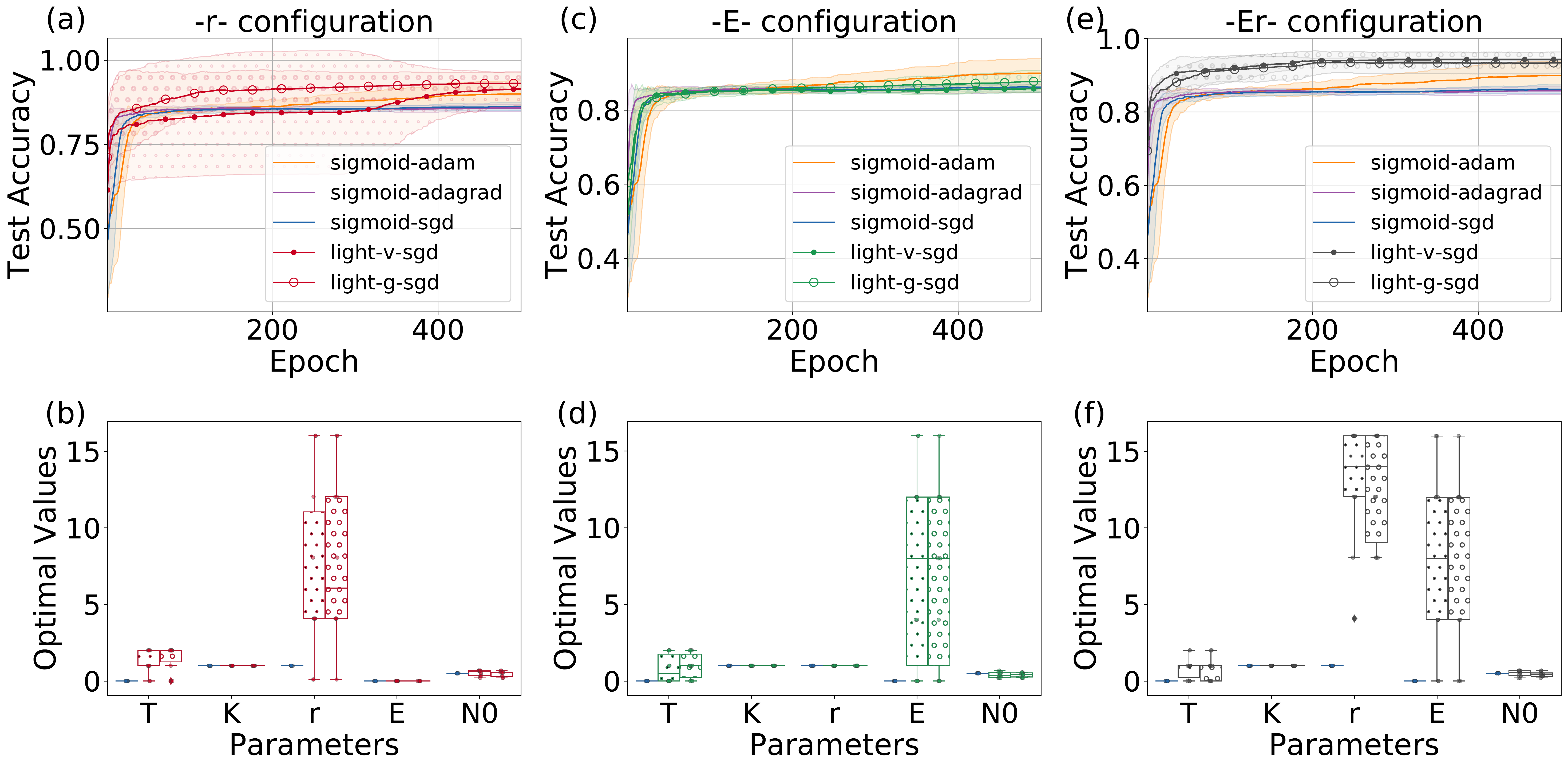}
	\caption{Test accuracy and optimal hyperparameters on MOONS (higher variance) for $L = 1$, $d_l = 5$, $m = 1000$, $n = 2$, \emph{noise} = 0.25}
	\label{fig::moonsL1std25}
\end{figure}

\newpage
\bibliographystyle{IEEEtran}
\bibliography{IEEEabrv,lightbib}

\begin{IEEEbiography}[{\includegraphics[width=1in,height=1.25in,clip,keepaspectratio]{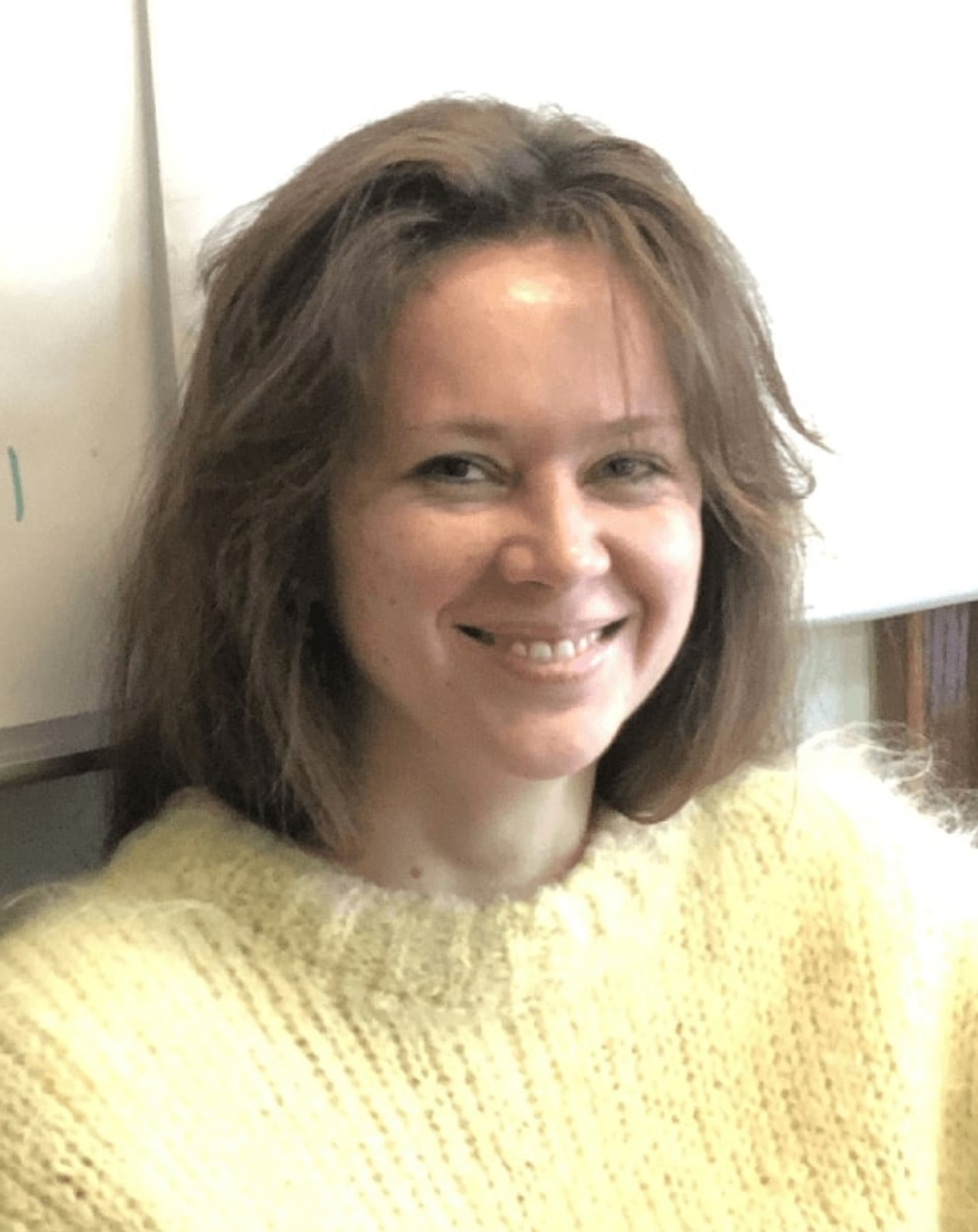}}]{Ilona Kulikovskikh}
	received her Candidate of Sciences (Ph.D.) in 2011 and Doctor of Sciences (Dr. Sc.) in 2020 from the Higher Attestation Committee in Moscow. In 2018-2019, she was a Visiting Scholar at the University of Zagreb and Ru\dj er Bo\v{s}kovi\'c Institute, where she worked on bio-inspired learning systems. She is currently with Samara University as a Full Professor and a Senior Researcher, leading projects and conducting transdisciplinary studies on explainable and reliable AI, evolutionary computation, and dynamic systems. She is a member of the Academy of Navigation and Motion Control.
	%a Full Professor and a Senior Researcher, giving lectures and leading transdisciplinary projects in academia and industry on 
	%explainable and reliable AI, evolutionary computation, and dynamic systems.
	%She is a member of the Academy of Navigation and Motion Control.
\end{IEEEbiography}

\begin{IEEEbiography}[{\includegraphics[width=1in,height=1.25in,clip,keepaspectratio]{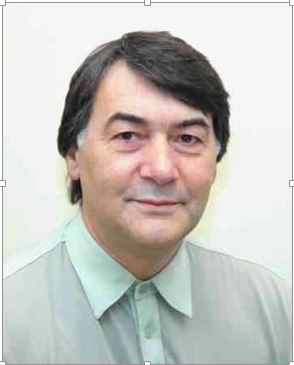}}]{Tarzan Legovi\'c}
	received his M.Sc. from the University of Toronto in 1976 and Ph.D. from the University of Zagreb in 1980.
	He serves as the president of the International Society for Ecological Modelling. 
	He is employed as a professor at the Libertas International University and the president of the Scientific Council at the OIKON Ltd. At the OIKON he also heads the Data Science Laboratory which is concerned with methods of AI
	and their applications to ecology. As an external scientific advisor at the R. Bo\v{s}kovi\'c institute, he helps in the development of predictive models for environmental management.
\end{IEEEbiography}


%\vfill


\end{document}